\documentclass[journal]{IEEEtran}
\ifCLASSINFOpdf
\else
\fi
\usepackage{cite}
\usepackage{amsmath,amssymb,amsfonts}
\usepackage{algorithm} 
\usepackage{graphicx}
\usepackage{caption}
\usepackage{subcaption}
\usepackage{textcomp}
\usepackage{xcolor}
\usepackage{array}
\usepackage{textcomp}
\usepackage{stfloats}
\usepackage{url}
\usepackage{verbatim}
\usepackage{amsthm, bm, hyperref, mathrsfs, indentfirst, longtable}
\usepackage{svg}
\usepackage{longtable}
\usepackage{tabularx}
\usepackage{makecell}
\usepackage{booktabs,float}
\usepackage{bbm}
\usepackage{balance}
\usepackage{threeparttable}
\usepackage{multirow}
\usepackage{booktabs}

\begin{document}
\title{M\textsuperscript{2}BeamLLM: Multimodal Sensing-empowered mmWave Beam Prediction with Large Language Models
}

\author{
    Can Zheng, 
    Jiguang He,~\IEEEmembership{Senior Member,~IEEE,}
    Chung G. Kang,~\IEEEmembership{Senior Member,~IEEE,}
    Guofa Cai,~\IEEEmembership{Senior Member,~IEEE,} 
    Zitong Yu,~\IEEEmembership{Senior Member,~IEEE,}
    M\'erouane Debbah,~\IEEEmembership{Fellow,~IEEE} 
    \thanks{C. Zheng and C. G. Kang are with the Department of Electrical and Computer Engineering, Korea University, Seoul 02841, South Korea (e-mail: zc331\_@korea.ac.kr, ccgkang@korea.ac.kr).}
    \thanks{J. He and Z. Yu are with the School of Computing and Information Technology, Great Bay University, Dongguan Key Laboratory for Intelligence and Information Technology, and Great Bay Institute for Advanced Study (GBIAS), Dongguan 523000, China (e-mail: jiguang.he@gbu.edu.cn, zitong.yu@ieee.org).}%
    \thanks{G. Cai is with School of Information Engineering, Guangdong University of Technology, Guangzhou, China (e-mail: caiguofa2006@gdut.edu.cn).}
    \thanks{M. Debbah is with the Center for 6G Technology, Khalifa University of Science and Technology, Abu Dhabi, United Arab Emirates (e-mail: merouane.debbah@ku.ac.ae).}
}

\maketitle
\begin{abstract}
    This paper introduces a novel neural network framework called M\textsuperscript{2}BeamLLM for beam prediction in millimeter-wave (mmWave) massive multi-input multi-output (mMIMO) communication systems. M\textsuperscript{2}BeamLLM integrates multi-modal sensor data, including images, radar, LiDAR, and GPS, leveraging the powerful reasoning capabilities of large language models (LLMs) such as GPT-$2$ for beam prediction. By combining sensing data encoding, multimodal alignment and fusion, and supervised fine-tuning (SFT), M\textsuperscript{2}BeamLLM achieves significantly higher beam prediction accuracy and robustness, demonstrably outperforming traditional deep learning (DL) models in both standard and few-shot scenarios. Furthermore, its prediction performance consistently improves with increased diversity in sensing modalities. Our study provides an efficient and intelligent beam prediction solution for vehicle-to-infrastructure (V$2$I) mmWave communication systems.
\end{abstract}


\IEEEpeerreviewmaketitle

\section{Introduction}
    
    \IEEEPARstart{M}{illimeter}-wave (mmWave) frequency bands offer broader channel bandwidths compared to traditional sub-$6$ GHz bands, enabling higher capacity for communication systems. However, their high propagation loss introduces new challenges for system design. While high-gain narrow beams formed by large antenna arrays can effectively compensate for the high path loss, the resulting strong directional characteristics make the beams highly sensitive to user mobility and environmental obstructions \cite{mmWave}. This sensitivity can lead to a significant decline in the stability of mmWave connectivity. In this context, beam alignment (BA) becomes a crucial technology for determining the optimal/sub-optimal beamforming direction, playing a key role in establishing reliable communication links.

    In vehicle-to-everything (V$2$X) scenarios, the high mobility of user equipment (UE) imposes even more stringent requirements on BA mechanisms. Compared to general mobile scenarios, the rapid topology changes unique to mmWave V$2$X significantly increase the complexity of real-time alignment for beams. Frequent beam misalignment can lead not only to communication link interruptions but also to severe degradation of communication quality. Existing BA solutions rely on repeated beam training processes to determine the optimal beam direction, but the resource-intensive nature of this process can encroach upon time slot resources allocated for data transmission, resulting in reduced system throughput. Therefore, developing intelligent BA mechanisms that offer efficiency and robustness has become a key research direction for ensuring the reliability of mmWave V$2$X communications.

    The recent advancements in integrated sensing and communication (ISAC) have brought significant attention to sensing-empowered beam prediction. The optimal beam pair between the base station (BS) and UE is determined by their instantaneous spatial positions and the surrounding environment. Proactively utilizing UE positioning data alongside sensing data measured by the BS for beam prediction is a viable strategy to eliminate beam training overhead. Beam prediction based on historical sensing data essentially involves analyzing the complex interplay between the temporal dimension (historical state evolution) and the spatial dimension (environmental physical characteristics) of sensing data. This analysis is aimed at constructing mathematical models that reveal hidden dynamic patterns and establish a dynamic mapping between sensing data and future beam states. This task is particularly well-suited for deep learning (DL) methods, which can effectively capture and model such intricate relationships \cite{ml}. Prior studies have explored the use of various communication and sensing modalities—such as sub-$6$ GHz channel state information (CSI) \cite{csi}, RGB images \cite{CV}, radar \cite{Radar}, LiDAR \cite{LiDAR}, and GPS \cite{GPS}—in conjunction with DL architectures for beam prediction.

    The rapid development of large language models (LLMs), such as ChatGPT \cite{GPT-4} and DeepSeek \cite{deepseek}, introduces new possibilities for physical-layer applications that go beyond the limitations of conventional methods. LLMs, with their billions of parameters, exhibit exceptional capabilities in natural language understanding and logical reasoning. Owing to extensive pretraining on diverse and large-scale datasets, LLMs can achieve strong performance with minimal task-specific fine-tuning. Compared to training ransformer models from scratch, LLMs require significantly less labeled data, making them particularly advantageous in real-world applications where collecting and annotating large-scale supervised datasets is often impractical. Recent efforts have explored leveraging LLMs for physical-layer tasks, such as mmWave beam prediction. A representative example is BeamLLM~\cite{BeamLLM}, which utilizes LLMs to predict beam directions in vehicle-to-infrastructure (V$2$I) scenarios. While BeamLLM demonstrates strong prediction performance, its reliance on a single sensing modality-namely RGB images-limits its adaptability in complex and dynamic environments where visual information alone may be insufficient due to occlusions, lighting changes, or adverse weather conditions. To address this limitation, a more recent study~\cite{CV+GPS} has attempted to integrate GPS positioning data and RGB images for multimodal beam prediction. This work highlights the potential of cross-modal fusion in enhancing robustness. However, it suffers from two limitations. First, it relies on large-scale GPT-$4$ models whose inference latency renders the approach unsuitable for real-time applications. For practical deployment, a more feasible solution would leverage historical data to predict future beam, aligning better with real-world requirements. Second, still only a limited number of modalities and lacking adaptability to environmental variations.
    
    Beyond this, some prior works have investigated more generalized multimodal fusion frameworks to improve prediction robustness. For example, Shi \textit{et al.} \cite{Multi_modal} explored combining multimodal sensing data, yet the fusion was static and based on naive concatenation. In contrast, Zhang \textit{et al.} \cite{MoE} introduced a more dynamic mixture-of-experts (MoE) framework, in which a gating network assigned weights to each modality. In addition, Cui \textit{et al.} \cite{Transf} established the relationship between different modalities through cross-attention.
    
    In this paper, we introduce M\textsuperscript{2}BeamLLM, a multimodal sensing empowered beam prediction framework built upon LLMs. Our major contributions are summarized as follows:
    \begin{enumerate}
        \item Our M\textsuperscript{2}BeamLLM framework integrates multiple data sources—images, radar, LiDAR, and GPS. This multimodal approach enables a more comprehensive understanding of the environment, significantly enhancing the accuracy and robustness of beam prediction, particularly in dynamic scenarios.
        \item We pioneer the application of LLMs, specifically GPT-$2$, to the beam prediction task, which has traditionally been addressed using DL architectures, e.g., recurrent neural networks (RNNs) \cite{CV}. By leveraging the superior generalization and reasoning capabilities of LLMs, we effectively model complex relationships within multimodal data, an approach that remains largely unexplored in existing studies.
        \item Through supervised fine-tuning (SFT) of a pretrained LLM \cite{SFT}, we enable the model to adapt to the beam prediction task with minimal training data and computational cost. This contrasts sharply with the resource-intensive approaches of training models from scratch, offering a more efficient solution for practical deployment.
        \item Experimental results demonstrate that M\textsuperscript{2}BeamLLM outperforms state-of-the-art DL models in both standard and few-shot prediction scenarios. Our framework achieves a Top-$1$ accuracy of $68.9\%$, surpassing the next-best method by $13.9\%$, and maintains superior performance even under challenging few-shot conditions.
        \item Through ablation studies, we provide empirical evidence on the impact of different sensing modality combinations on prediction performance. This analysis reveals that modality diversity generally enhances prediction accuracy, offering new insights into the value of multimodal fusion in beam prediction and advancing the understanding of multi-source data contributions in this domain.
    \end{enumerate}
    
    The rest of this paper is organized as follows: Section II establishes the system model and problem formulation of V$2$I mmWave beam prediction. The proposed M\textsuperscript{2}BeamLLM framework is presented in Section III. Section IV presents extensive simulation results, including performance comparisons with other methods, along with some ablation studies. Finally, we conclude our work in Section V.

    \textit{Notations}: Bold lowercase letters denote vectors (e.g., $\mathbf{x}$), and bold uppercase letters denote matrices (e.g., $\mathbf{X}$). The superscripts $(\cdot)^\mathsf{T}$ and $(\cdot)^\mathsf{H}$ represent the transpose and Hermitian (conjugate transpose) operations, respectively. The symbol $\oslash$ denotes element-wise division. The operator $\mathbb{E}[\cdot]$ denotes the statistical expectation, while $|\cdot|_2$ denotes the Euclidean norm of a vector, and $|\cdot|$ returns the magnitude of a complex number. The functions $\min(\cdot)$ and $\max(\cdot)$ return the minimum and maximum element, respectively. The indicator function $\mathbbm{1}\{\cdot\}$ equals $1$ if the condition inside the braces is true, and $0$ otherwise. Unless otherwise specified, $\mathbb{R}$ and $\mathbb{C}$ denote the sets of real and complex numbers, respectively.

    \begin{figure}[ht]
        \centering
        \includegraphics[width=\linewidth]{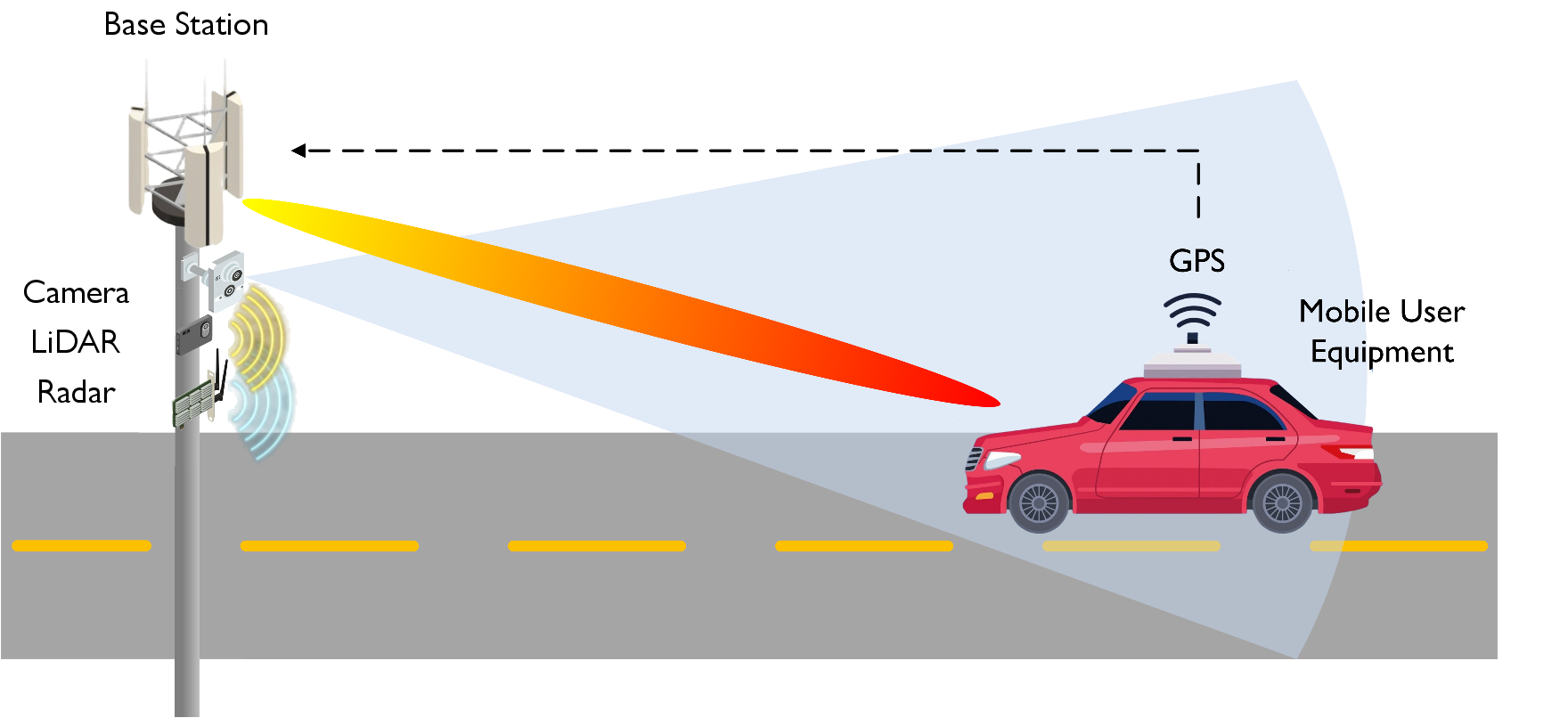}
        \caption{Illustration of the V$2$I system model: The BS is equipped with a camera, radar, and LiDAR, while a GPS-RTK system provides the UE's localization information to the BS.}
        \label{fig:multimodal}
    \end{figure}
    
\section{System Model And Problem Formulation}
    \subsection{System Model}
    As shown in Fig. \ref{fig:multimodal}, without loss of generality, we consider a wireless communication system with $N$-antenna BS and a single-antenna UE, employing a pre-defined beamforming codebook denoted as $\mathcal{F}=\{\mathbf{f}_m\}_{m=1}^M$, where $\mathbf{f}_m \in \mathbb{C}^{N\times 1}$ represents the $m$-th beamforming vector in the codebook. At time instance $t$, the transmitter selects a beam $\mathbf{f}_{m[t]}\in \mathcal{F}$ for signal transmission, where $m[t]$ denotes the index of the beamforming vector at time instant $t$. The received signal is modeled as
    \begin{align}
        y[t] = \mathbf{h}^\mathsf{H}[t]\mathbf{f}_{m[t]}s + n[t].
    \end{align}
    where $\mathbf{h}[t]$ is the mmWave channel vector, $s$ is the unit-power transmitted symbol, and $n[t]\sim\mathcal{CN}(0,\sigma_n^2)$ is complex additive Gaussian noise. The optimal beam selection strategy maximizes the effective channel gain as follows:
    \begin{align}
        \mathbf{f}_{m^{*}[t]} = \arg \max_{m\in\{1,\cdots,M\}} \left|\mathbf{h}^\mathsf{H}[t]\mathbf{f}_{m[t]}\right|^2,
    \end{align}
    where $m^{*}[t]$ denotes the index of the optimal beamforming vector at time instant $t$. This selection criterion ensures the highest possible received signal-to-noise ratio (SNR) under the given codebook constraints.

    \begin{figure*}[ht]
        \centering
        \includegraphics[width=\textwidth]{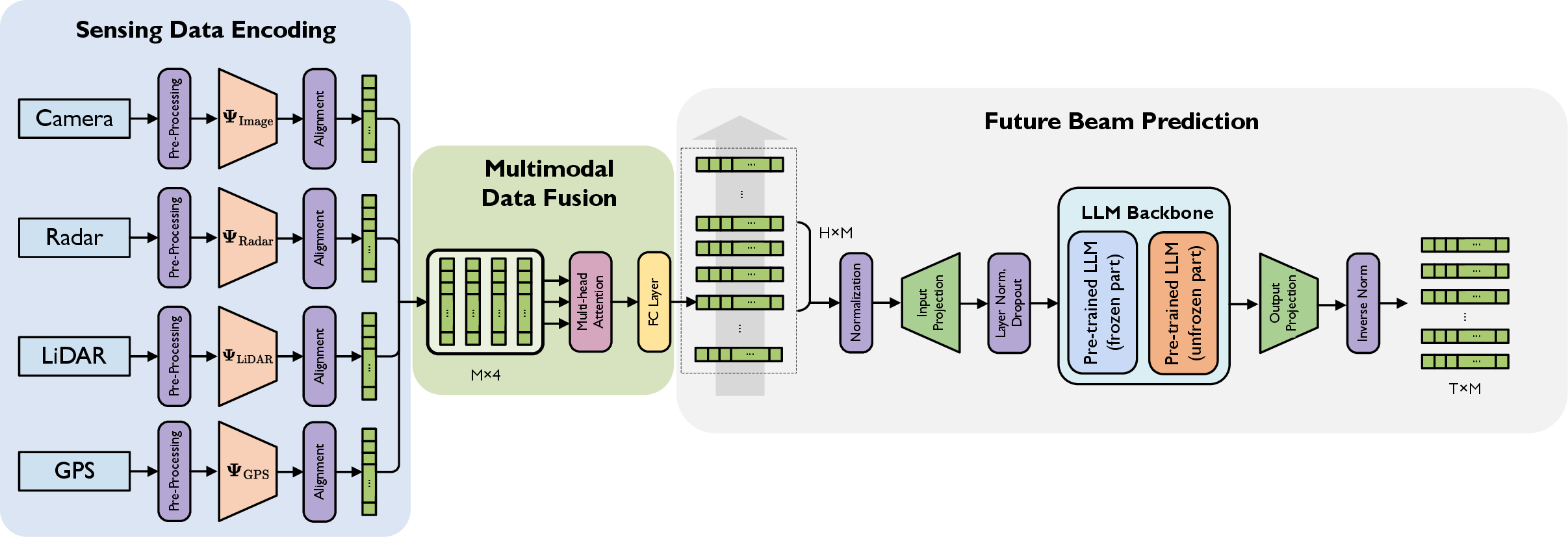}
        \caption{The model framework of M\textsuperscript{2}BeamLLM. In the sensing data encoding part, it encodes and aligns multimodal sensing data from camera, radar, LiDAR, and GPS. It then fuses these diverse inputs using a multi-head attention mechanism. Subsequently, the fused data undergoes input projection before being fed into an LLM backbone, which includes both frozen and unfrozen pre-trained components. Finally, an output projection and inverse normalization are applied to predict future beams.}
        \label{fig:MMBeamLLM}
    \end{figure*}
    
    \subsection{Problem Formulation}
    We aim to design a beam prediction strategy that maximizes the probability of correctly identifying the optimal beam index $m^{*}[t]$ over a future horizon $t=1, 2, \cdots,T$, leveraging historical observations from time slots $\tau=-H+1,-H+2\cdots,0$. The performance metric is defined as the expected accuracy over $T$:
    \begin{align}
        \mathcal{P} = \mathbb{E}\left[\frac{1}{T}\sum_{t=1}^T \mathbbm{1}{\left\{\hat{m}[t] = m^{*}[t]\right\}}\right],
    \end{align}
    where the expectation is taken over the joint distribution of channel realizations $\{\mathbf{h}[t]\}_{t=1}^T$ and historical observations $\{\mathbf{h}[t]\}_{\tau=-H+1}^0$. While this formulation relies on temporal correlations in channel realizations, such methods face the following fundamental limitations:
    \begin{itemize}
        \item \textbf{Channel dynamics dependency:} Traditional approaches assume predictable temporal patterns (e.g., slow fading, Markovian transitions), which collapse in non-stationary scenarios (e.g., sudden blockage by vehicles, UAV swarms maneuvering).
        \item \textbf{Reactive adaptation:} Channel measurements only provide posterior information, delaying beam adjustments until after link degradation occurs.
    \end{itemize}
    
    In contrast, sensing data enables proactive prediction by directly observing the physical causes of channel variations. This paradigm shift replaces statistical channel extrapolation with geometric-environmental reasoning, achieving robustness against abrupt channel dynamics through environment-aware prediction.

    \section{M\textsuperscript{2}BeamLLM Framework}
    In this section, we introduce M\textsuperscript{2}BeamLLM to tackle the multimodal sensing data empowered beam prediction task outlined in Section II. The general M\textsuperscript{2}BeamLLM framework is shown in Fig. \ref{fig:MMBeamLLM}, which consists of the following key components: sensing data encoding, multimodal data fusion, and future beam prediction.

    \subsection{Multimodal Feature Encoding}

    The BS is equipped with a set of sensing modalities denoted by $\Omega = \{\text{Image}, \text{Radar}, \text{LiDAR}, \text{GPS}\}$. For each modality $\omega \in \Omega$, the raw sensing data at time $t$ is represented as $\mathbf{X}_\omega[t]$, with the following specifications:
    \begin{itemize}
        \item \textbf{Image Data}: $\mathbf{X}_\text{Image}[t]\in \mathbb{R}^{W_I\times H_I \times C_I}$, where $W_I$, $H_I$ and $C_I$ represent the spatial width, height, and RGB/IR spectral channels, respectively.
        \item \textbf{Radar Data}: $\mathbf{X}_\text{Radar}[t] \in \mathbb{C}^{M_R \times S_R \times A_R}$, where the dimensions correspond to antenna, sampling, and chirp, providing information on the target's angle, distance, and velocity.
        \item \textbf{LiDAR Data}: $\mathbf{X}_\text{LiDAR}[t] \in \mathbb{R}^{N_L[t]\times 3}$, with $N_L[t]$ being the time-varying number of $3$D points $(x,y,z)$ in the LiDAR point cloud.
        \item \textbf{GPS Data}: $\mathbf{X}_\text{GPS}[t] \in \mathbb{R}^{2\times 1}$, containing the latitude and longitude coordinates of the UE. Although GPS data incurs uplink latency from UE to the BS, its computational simplicity compared to other latency-intensive modalities enables synchronization via buffer delay or some other schemes.
    \end{itemize}
    
    We formalize the feature extraction process for each sensing modality, where $\mathbf{\Psi}_\omega(\cdot)$ denotes the modality-specific feature encoder for $\omega \in \Omega$. Their specific architecture is shown in Fig. \ref{fig:Encoders}. A detailed description is given below:

        \subsubsection{Image Data Feature Extraction} The image feature extraction pipeline processes raw input $\mathbf{X}_\text{Image}[t]$ through three stages: standardization, backbone processing, and dimension compression. First, each image is reshaped to $224\times224\times3$ and normalized using ImageNet parameters \cite{ImageNet}:
        \begin{align}
            \mathbf{x}_\text{Image}^\text{Norm} = (\text{Reshape}(\mathbf{X}_\text{Image}[t])-\boldsymbol{\mu}_\text{Image}) \oslash \boldsymbol{\sigma}_\text{Image},
        \end{align}
        where $\text{Reshape}(\cdot)$ represents image reshaping operation, $\boldsymbol{\mu}_\text{Image}$ and $\boldsymbol{\sigma}_\text{Image}$ are the mean and standard deviation of the ImageNet dataset, respectively.
        
        The normalized tensor is then fed into a pretrained ResNet-$18$ backbone \cite{ResNet} (with removed classification layers), followed by a learnable fully connected (FC) layer that compresses features to $M$ dimensions:
        \begin{align}
            \tilde{\mathbf{x}}_\text{Image}[t]=\mathbf{\Psi}_\text{Image}\left(\mathbf{x}_\text{Image}^\text{Norm}\right) \in \mathbb{R}^{M}.
        \end{align}

        \begin{figure*}[!t]
        \centering
          \begin{subfigure}{0.9\textwidth}
            \includegraphics[width=0.727\textwidth]{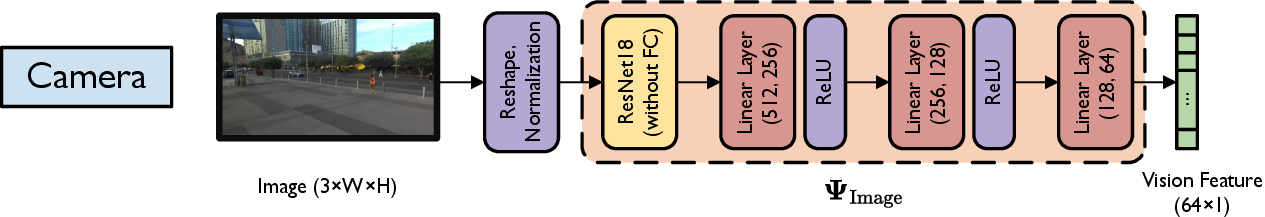}
            \caption{Vision feature encoder. It generates visual features  by preprocessing the input image, ResNet-$18$ feature extraction and multilayer linear transformation with ReLU activation.}
            \label{fig:vision_encoder}
          \end{subfigure}
          \vspace{0.5em}
          \begin{subfigure}{0.9\textwidth}
            \includegraphics[width=0.805\textwidth]{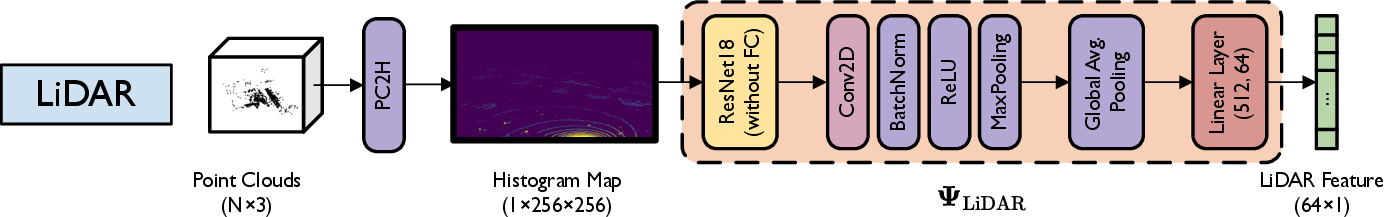}
            \caption{LiDAR feature encoder. It converts LiDAR point cloud data into histograms and then extracts LiDAR features through ResNet-$18$, convolutional layers, pooling, and linear layers.}
            \label{fig:lidar_encoder}
          \end{subfigure}
          \vspace{0.5em}
          \begin{subfigure}{0.9\textwidth}
            \includegraphics[width=\textwidth]{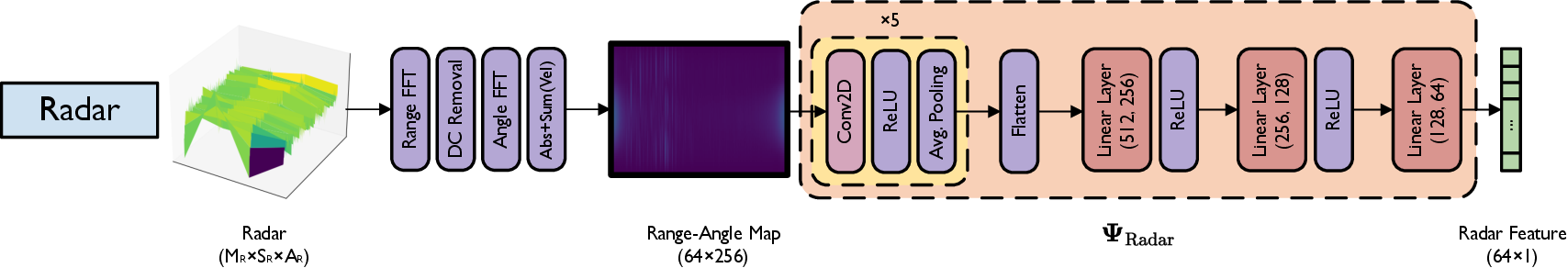}
            \caption{Radar feature encoder. It processes the raw radar data with Range FFT, Angle FFT, to generate Range-Angle Map, and then extracts the radar features through a series of operations such as convolution, pooling, and linear layer.}
            \label{fig:radar_encoder}
          \end{subfigure}
          \vspace{0.5em}
          \begin{subfigure}{0.9\textwidth}
            \includegraphics[width= 0.64\textwidth]{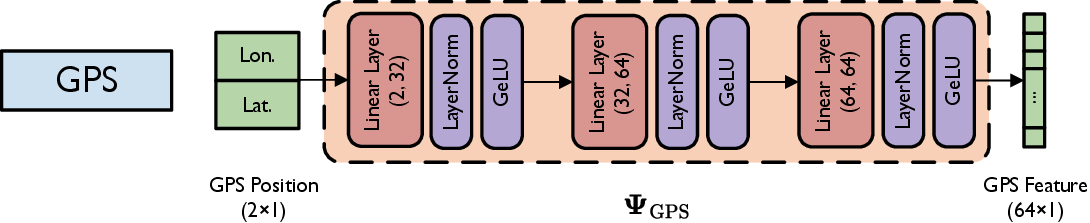}
            \caption{GPS feature encoder. It processes the longitude and latitude information through a simple MLP to extract GPS features.}
            \label{fig:gps_encoder}
          \end{subfigure}
    
          \caption{Multimodal data encoding module for the M\textsuperscript{2}BeamLLM system. This module demonstrates in detail the independent feature encoding process for (a) camera, (b) LiDAR, (c) radar, and (d) GPS sensors, converting the raw sensor data into a unified feature representation for subsequent multimodal data fusion.}
          \label{fig:Encoders}
        \end{figure*}
        
        \subsubsection{Radar Data Feature Extraction} The radar feature extraction pipeline begins by processing raw radar data $\mathbf{X}_\text{Radar}[t]$ through a two-dimensional fast Fourier transform ($2$D-FFT) applied to each chirp signal along the antenna and fast-time dimensions. This transformation generates a range-angle (RA) map:
        \begin{align}           
            {\mathbf{X}}_\text{Radar}^\text{RA}[t] &= \sum_{a=1}^{A_R} |\text{FFT}_{2\text{D}}\left(\mathbf{X}_\text{Radar}[t][:,:,a] \right)|\in \mathbb{R}^{M_F \times S_R},
        \end{align}
        where $\text{FFT}_{2\text{D}}(\cdot)$ represents the $2$D-FFT operation. The antenna dimension is zero-padded from $M_R$ to $M_F$ ($M_F > M_R$) to achieve angular oversampling, reducing the effective angular sampling interval and enhancing spectral detail resolution \cite{ZP_FFT}. The resulting RA map is then encoded into a feature vector through a convolutional neural network \cite{LeNet}:
        \begin{align}
            \tilde{\mathbf{x}}_\text{Radar}[t]&=\mathbf{\Psi}_\text{Radar}\left({\mathbf{X}}_\text{Radar}^\text{RA}[t]\right)\in \mathbb{R}^{M}.
        \end{align}
        \subsubsection{LiDAR Data Feature Extraction} The LiDAR feature extraction pipeline processes raw point cloud data $\mathbf{X}_\text{LiDAR}[t]$ through geometric transformation and neural encoding. First, the $3$D points are projected onto a $256\times 256$ $2$D grid to create a histogram representation $\mathbf{X}_\text{LiDAR}^\text{Hist}[t]$, where point counts per cell are clipped to $5$ and normalized to $[0,1]$ for robustness:
        \begin{align}
        \mathbf{X}_\text{LiDAR}^\text{Hist}[t] &=\text{PC2H}\left(\mathbf{X}_\text{LiDAR}[t]\right)\in \mathbb{R}^{1\times 256\times 256},
        \end{align}
        where $\text{PC2H}(\cdot)$ denotes the point-cloud-to-histogram transformation. 
        This single-channel feature map is then encoded into an $M$-dimensional feature vector using a modified ResNet-$18$ architecture:

        \begin{align}
            \tilde{\mathbf{x}}_\text{LiDAR}[t] = \mathbf{\Psi}_\text{LiDAR}\left(\mathbf{X}_\text{LiDAR}^\text{Hist}[t]\right)\in \mathbb{R}^{M},
        \end{align}
    
        \subsubsection{GPS Data Feature Extraction} The GPS feature extraction process begins by normalizing raw coordinate data $\mathbf{X}_\text{GPS}$ to ensure compatibility with subsequent fusion operations. Using min-max scaling applied across the entire historical dataset $\mathbf{\mathbf{X}_\text{GPS}}\in \mathbb{R}^{2\times H}$, we maintain temporal consistency through global normalization:
        \begin{align}
            \mathbf{X}_\text{GPS}^\text{Norm}[t] = \frac{\mathbf{X}_\text{GPS}[t] - \text{min}(\mathbf{\mathbf{X}_\text{GPS}})}{\text{max}(\mathbf{\mathbf{X}_\text{GPS}}) - \text{min}(\mathbf{\mathbf{X}_\text{GPS}})}.
        \end{align}
        
        The normalized coordinates are then mapped to an $M$-dimensional feature space through a multilayer perceptron (MLP) network:
        \begin{align}
            \tilde{\mathbf{x}}_\text{GPS}[t] = \mathbf{\Psi}_\text{GPS}\left(\mathbf{X}_\text{GPS}^\text{Norm}[t]\right) \in \mathbb{R}^{M}.
        \end{align}
        \subsection{Multimodal Alignment}
        Multimodal alignment seeks to bridge the semantic discrepancies among heterogeneous sensing modalities by projecting modality-specific features into a shared embedding space, thereby enabling effective cross-modal fusion \cite{Alignment}. In this work, we employ a contrastive learning-based alignment strategy inspired by CLIP \cite{CLIP}, which enforces geometric consistency among modalities through cosine similarity constraints. Specifically, the feature vector $\tilde{\mathbf{x}}_\omega[t]$ extracted from each modality $\omega \in \Omega$ at time step $t$ is first $\ell_2$-normalized:
        \begin{align}
            \bar{\mathbf{x}}_\omega[t]&=\frac{\tilde{\mathbf{x}}_\omega[t]}{\Vert \tilde{\mathbf{x}}_\omega[t]\Vert_2},\quad \forall \omega\in\Omega,
        \end{align}
        projecting all features onto the unit hypersphere $\mathbb{S}^{M-1}$. This normalization facilitates stable optimization using cosine similarity as the alignment metric. Cross-modal alignment is then enforced by maximizing the pairwise cosine similarity between normalized feature vectors. These similarities are organized into a symmetric matrix $\mathbf{S}[t] \in [-1, 1]^{|\Omega| \times |\Omega|}$, where each element is computed as
        \begin{align}
            S_{\omega_1,\omega_2}[t] = \bar{\mathbf{x}}_{\omega_1}[t]\bar{\mathbf{x}}^{\top}_{\omega_2}[t], \quad \forall \omega_1, \omega_2 \in \Omega
        \end{align}
        which quantitatively reflects the semantic consistency between modalities $\omega_1$ and $\omega_2$ at time $t$.

        By encouraging high cross-modal similarity, this geometric constraint implicitly disentangles modality-invariant semantics from modality-specific noise, thereby enabling robust and generalizable feature fusion—especially under distributional shifts or missing modalities.

        \subsection{Multimodal Feature Fusion}
        To integrate features from heterogeneous modalities, we design a transformer-based fusion module that captures cross-modal dependencies through self-attention mechanisms. For each time step $t$, the normalized features $\{\bar{\mathbf{x}}_\omega[t]\}_{\omega \in \Omega}$ are stacked into a matrix $\mathbf{F}[t] = [\bar{\mathbf{x}}^\mathsf{T}_\text{Image}[t], \bar{\mathbf{x}}^\mathsf{T}_\text{Radar}[t], \bar{\mathbf{x}}_\text{LiDAR}^\mathsf{T}[t], \bar{\mathbf{x}}_\text{GPS}^\mathsf{T}[t]]^\mathsf{T}\in \mathbb{R}^{M\times 4}$, where each row corresponds to a modality. To model inter-modal interactions, we apply a multi-head self-attention operation, where the queries, keys, and values are all set to $\mathbf{F}^\mathsf{T}[t]$, i.e., $Q(t) = K(t) = V(t) = \mathbf{F}^\mathsf{T}[t]$:
        \begin{align}
            \mathbf{A}[t] = \text{MultiHead}(\mathbf{F}^\mathsf{T}[t], \mathbf{F}^\mathsf{T}[t], \mathbf{F}^\mathsf{T}[t]) \in \mathbb{R}^{|\Omega|\times M}.
        \end{align}
        
        Each attention head captures distinct correlation patterns among modalities, and the outputs are concatenated and projected back to the original dimension $M$. This enables dynamic weighting and context-aware fusion of modality-specific features. The modality-aware representations are then aggregated across modalities by summing the attended vectors:
        \begin{align}
            \mathbf{z}[t] = \text{FFN}\left( \sum_{\omega \in \Omega} \mathbf{A}[t]_{(\omega,:)}\right) \in \mathbb{R}^{M},
        \end{align}
        where $\text{FFN}(\cdot)$ denotes a position-wise feed-forward network (FFN) that refines the fused representation. Finally, the time-series feature sequence is formed by concatenating the fused embeddings over a historical window of length $H$:
        \begin{align}
            \mathbf{Z} = [\mathbf{z}[-H+1], \mathbf{z}[-H+2], \cdots, \mathbf{z}[0]]^\mathsf{T} \in \mathbb{R}^{H\times M},
        \end{align}
        which serves as the input to downstream temporal modeling modules for beam prediction.

        \subsection{Prediction}
        The fused sequence obtained from the previous steps is fed into a prediction network  $\text{Pred}(\cdot)$ to make predictions, as follows:
        \begin{align}
            \hat{\mathbf{P}} = \text{Pred}(\mathbf{Z}) \in \mathbb{R}^{T\times M}.
        \end{align}
        
        Finally, for each time sample $t$,  we can choose the optimal beam index for prediction:
        \begin{align}
            \hat{m}[t] = \arg \max_{m} \mathbf{\hat{\mathbf{P}}}[t,m].
        \end{align}

    \subsection{Model Structure}
    \textbf{Input Projection.} 
    Given an input feature sequence $\mathbf{Z}$, the input embedding module is designed to project $\mathbf{Z}$ into a $d_\text{LLM}$-dimensional embedding space, effectively treating each time step as a token embedding suitable for LLMs. In other words, this process enables LLMs to ``comprehend'' the input feature sequence. Prior to embedding, $\mathbf{Z}$ is normalized to have zero mean and unit variance, which stabilizes training and preserves the relative relationships between features. The normalized sequence is then linearly transformed into variable-length embeddings compatible with the LLM input format.

    \textbf{Pretrained LLM and SFT.}
    We employ a pretrained LLM backbone, freezing most of its self-attention and feed-forward layers during fine-tuning to retain the general representations learned from large-scale data. To enable task adaptation, the parameters of the last few layers are unfrozen and updated. This selective fine-tuning strategy balances model stability and plasticity: it mitigates overfitting risks on relatively small datasets while allowing higher layers to specialize in task-specific refinements without degrading the foundational knowledge captured in lower layers. In M\textsuperscript{2}BeamLLM, this strategy is applied to beam prediction via SFT, requiring only a small amount of labeled data from the target task. By freezing the majority of layers, our approach efficiently adapts the pretrained LLM to the specific demands of beam prediction in mmWave systems, while significantly reducing the training cost.
    
    \textbf{Output Projection.}
    The contextualized embeddings output by the LLM are projected back to the original feature dimension $M$ through a learnable linear layer, producing the final predictions aligned with the input feature space.

    \subsection{Learning Phase}
    The training process consists of two parts: encoder pre-training and beam prediction module fine-tuning. The encoder pre-training simply takes the one-hot encoding vector corresponding to the modal input with the optimal beam index as a supervised learning pair and then performs $M$ classification with the following loss function:
    \begin{align}
        \mathcal{L}_\text{enc} = -\sum_{m=1}^{M}p_m\log(\sigma_\text{Softmax}(\hat{p}_m)),
    \end{align}
    where $p_m\in \{0,1\}$ denotes the ground-truth one-hot encoded optimal beam index, $\hat{p}_m$ represents the model prediction probability for the $m$-th beam, and $\sigma_\text{Softmax}(\cdot)$ is the softmax activation function.
    
    For beam prediction module fine-tuning part, the composite loss function comprises two key objectives:
    \subsubsection{Prediction Task Objective} We optimize beam index prediction accuracy through cross-entropy loss defined as:
    \begin{align}
        \mathcal{L}_1 = -\frac{1}{T}\sum_{t=1}^{T} \sum_{m=1}^{M}p_m[t]\log(\sigma_\text{Softmax}(\hat{p}_m[t])).
    \end{align}

    \subsubsection{Multimodal Alignment Objective} Inter-modal feature similarity is enforced via normalized temperature-scaled contrastive loss defined as:
    \begin{align}
        \mathcal{L}_2 &= -\frac{\sum_{\tau=-H+1}^{0} \sum_{\omega_1,\omega_2\in \Omega}p_m[t]\log\left(\sigma_\text{Softmax}\left(\frac{S_{\omega_1,\omega_2}[t]}{\alpha}\right)\right)}{H\cdot|\Omega|(|\Omega|-1)},
    \end{align}
    where temperature parameter $\alpha>0$ modulates similarity concentration: smaller $\alpha$ sharpens hard negative mining while larger $\alpha$ promotes softer distribution.

    \subsubsection{Composite Optimization Objective}
    The unified training loss is defined by combining both objectives with weight $\lambda \in [0,+\infty)$ as follows:
    \begin{align}
    \mathcal{L} = \mathcal{L}_1 + \lambda\mathcal{L}_2.
    \end{align}

\section{Simulation Results}
    
    \subsection{Experimental Settings} 
    \subsubsection{Dataset Processing}
    In this study, we employ the DeepSense $6$G dataset \cite{DeepSense6G} to simulate a V$2$I mmWave communication environment. In this configuration, a stationary BS is equipped with an array of sensing technologies, including an RGB camera, radar, and LiDAR sensor, which collectively interact with a mobile UE functioning as a mmWave transmitter. The UE incorporates a GPS-RTK system that continuously streams meter-accurate positional data to the BS.
    
    The dataset is partitioned into training ($70\%$, $2,138$ samples) and testing ($30\%$, $917$ samples) sets, where each sample represents a complete vehicle pass-by event comprising synchronized multimodal sensing data sequences and beam index sequences. We decompose each sequence into input-output pairs using a $13$-frame sliding window, configuring two prediction tasks: 1) standard prediction with $H=8$ historical observations and $T=5$ future beams, and 2) few-shot prediction with $H=3$ and $T=10$, where the model processes multimodal sensing data inputs $\{\tilde{\mathbf{x}}_\text{Image}[t], \tilde{\mathbf{x}}_\text{LiDAR}[t],\tilde{\mathbf{x}}_\text{Radar}[t],\tilde{\mathbf{x}}_\text{GPS}[t]\}_{\tau=-H+1}^0$ while learning illumination-robust environment-beam correlations. 
    
    \subsubsection{Baselines}
    The default LLM backbone used in M\textsuperscript{2}BeamLLM is the distilled GPT-$2$ model \cite{GPT2}. As a longstanding benchmark and representative of encoder-based architectures, BERT is selected as the primary comparison baseline.\footnote{Both models are publicly available on the Hugging Face, with the respective URLs as follows: https://huggingface.co/tanyildizderya/distilled-gpt2 and https://huggingface.co/google-bert/bert-base-uncased.} We compare our approach with several time-series forecasting models, including RNN-based model (GRU \cite{GRU}, LSTM \cite{LSTM}), linear model (NLinear \cite{NLinear}), and transformer-based model (Informer \cite{Informer}).
    
    \subsubsection{Implementation Settings}

    Our experiments are conducted on a single NVIDIA A$100$-$40$ GB GPU via Google Colab, implemented in PyTorch. We employ the Adam optimizer \cite{ADAM} with an initial learning rate of $10^{-3}$, which decays by a factor of $0.5$ every $5$ epochs, alongside a batch size of $16$ and a training duration over $30$ epochs. The remaining simulation parameters are listed in Table \ref{tab:param}.

    \begin{table}[t]
    \centering
    \caption{Default Parameter Settings.}
    \label{tab:param}
    \begin{tabular}{|c|c|}  
    \hline  
    \textbf{Parameters} & \textbf{Value} \\ 
    \hline 
    Scenario   & DeegSense $6$G Scenario $32$ \\
    \hline
    $W_I, H_I, C_I$ & $960, 540, 3$ \\
    \hline
    $M_R, S_R, A_R$ & $4,256,128$\\
    \hline
    $N_L$ & $\sim 18000$ \\
    \hline
    $\boldsymbol{\mu}_\text{Image}$ & $[0.485, 0.456, 0.406]$ \\
    \hline 
    $\boldsymbol{\sigma}_\text{Image}$ & $[0.229, 0.224, 0.225]$\\
    \hline
    $d_\text{LLM}$      & $768$           \\
    \hline 
    $\alpha$            & $0.07$          \\
    \hline  
    $\lambda$           & $1$             \\
    \hline  
    Number of unfrozen layers & $2$            \\
    \hline 
    \end{tabular}
    \end{table}
    
    \subsubsection{Performance Metrics}
    \begin{itemize}
        \item \textbf{Top-$K$ Accuracy:} It measures whether the true label is included among the top-$K$ predicted beamforming vectors with the highest probabilities, defined as:
        \begin{align}
            \text{Top-}K \text{ Accuracy}=\frac{1}{N_\text{Test}} \sum_{n=1}^{N_\text{Test}} \mathbbm{1}{\{m_n \in \mathcal{Q}_K\}},
        \end{align}
        where $N_\text{Test}$ represents the total number of samples in the test set, $m_n$ denotes the index of the ground truth optimal beam for the $n$-th sample, and $\mathcal{Q}_K$ is the set of indices for the top $K$ predicted beams, sorted by the element values in $\hat{\mathbf{P}}$ for each time sample.
        \item \textbf{Distance-Based Accuracy (DBA) Score:} It measures the accuracy of the distance between the predicted beam and the ground truth beam. For this, we utilize the top-$K$ predicted beams. The DBA-Score is defined as
        \begin{align}
            \text{DBA-Score} = \frac{1}{K} \sum_{k=1}^K Y_k,
        \end{align}
        where 
        \begin{align}
            Y_k = 1- \frac{1}{N_\text{Test}} \sum_{n-1}^{N_\text{Test}}\min_{1 \leq k' \leq k} \left[ \min \left( \frac{|m_n-\hat{m}_{n,'k}|}{\Delta},1\right)\right].
        \end{align}
        
        Here, $\hat{m}_{n,k'}$ represents the prediction for the $k$-th most-likely beam index of sample $n$, and $\Delta$ is a normalization factor. Unlike traditional top-$K$ accuracy  that focus on whether a prediction hits the target, it introduces a  ``precision-aware'' evaluation criterion that allows for distance errors. If the prediction is close but not entirely correct, it can still receive a favorable score.
    \end{itemize}

    To facilitate presentation, the two performance metrics in the simulation results are averaged over all predicted time sequences.

    \subsection{Dataset Analysis and Visualization}

    In this section, we present an analysis and visualization of the dataset used. For visual and GPS data, we visualize multiple samples due to the significant variation between different instances. In contrast, Radar and LiDAR data exhibit less variation across samples; therefore, we provide a single sample visualization as a representative example.

\begin{figure}[t]
        \centering
        \includegraphics[width=\linewidth]{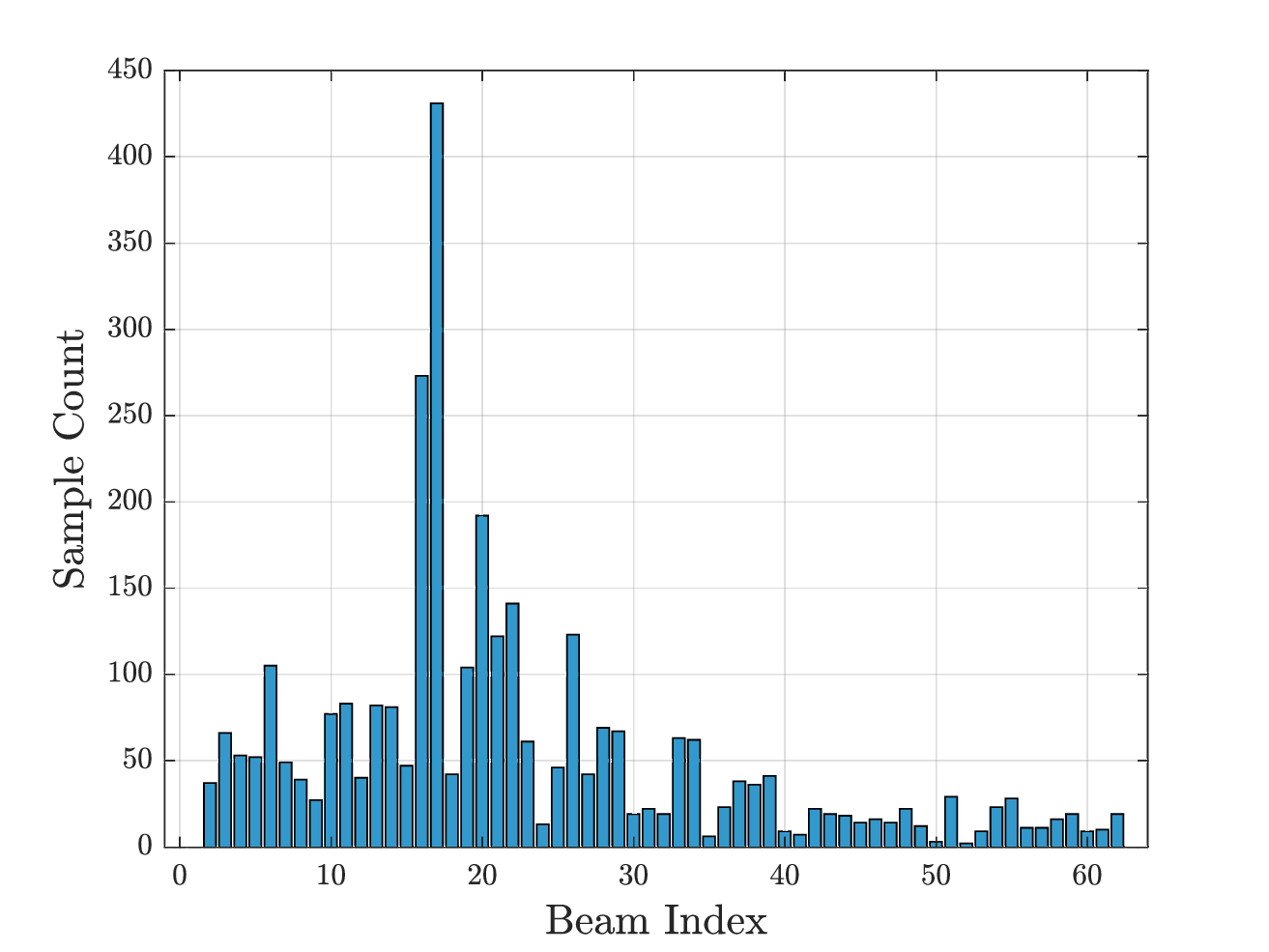}
        \caption{Optimal beam index frequency distribution.}
        \label{fig:freq}
    \end{figure}

    Fig. \ref{fig:freq} shows the frequency with which each beam index was optimal in the dataset in Scenario $32$. We observed that optimal beam indices in the range of $10$ to $30$ occur with higher frequency, with a particularly prominent peak around index $m^{*}=17$.

\begin{figure}[t]
        \centering
        \includegraphics[width=\linewidth]{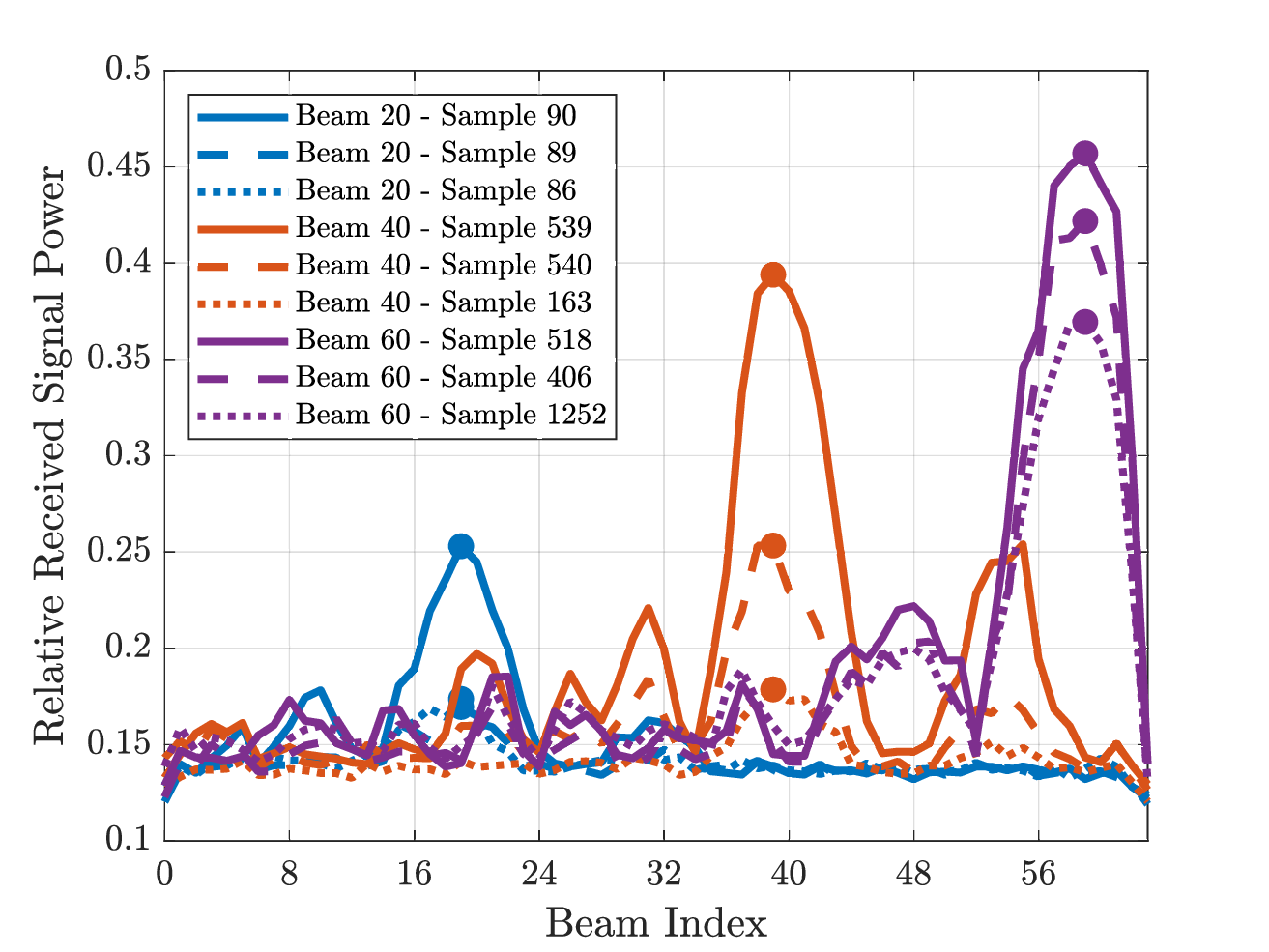}
        \caption{Comparison of power profiles for the first $3$ samples associated with optimal beams $20$, $40$, and $60$, respectively.}
        \label{fig:pow}
    \end{figure}

    Fig. \ref{fig:pow} illustrates the relative received power distribution across various beam indices ($20$, $40$, $60$) for the first three samples, highlighting how peak power varies with different beam selections. Note that due to the complexity of absolute power calibration, the DeepSense $6$G dataset provides dimensionless relative received signal power \cite{DeepSense6G}. From the figure, it can be seen that as the optimal beam index increases, the relative received power shows an overall increasing trend. With a beam index of $20$, the peak of the relative received power is around $0.25$; with a beam index of $40$, the peak of the relative received power improves to around $0.4$; and when the beam index is $60$, the relative received power reaches the highest level of around $0.45$. This implies that the distance between transceivers in the dataset may be smaller when the beam index is larger, leading to an increase in received power.

\begin{figure}[!t]
        \centering
          \begin{subfigure}{\linewidth}
          \centering
            \includegraphics[width=\textwidth]{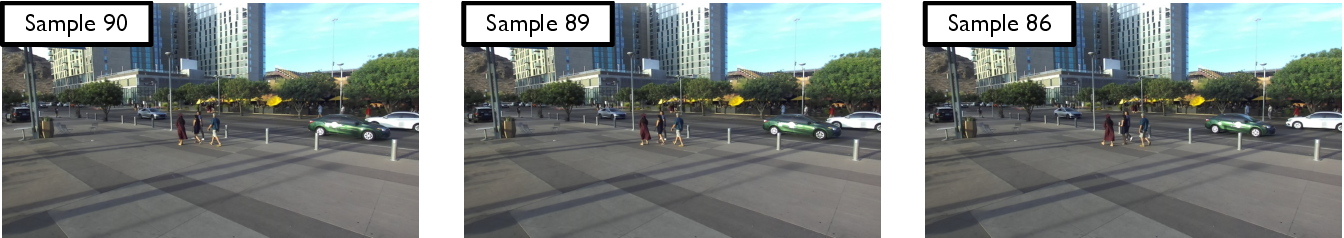}
            \caption{Vision data visualization for the first $3$ samples associated with optimal beam $20$.}
            \label{fig:20}
          \end{subfigure}
          \vspace{0.5em}
          \begin{subfigure}{\linewidth}
          \centering
            \includegraphics[width=\textwidth]{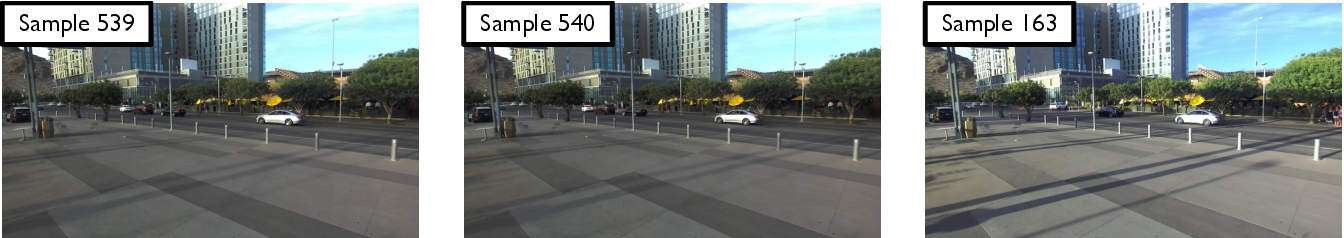}
            \caption{Vision data visualization for the first $3$ samples associated with optimal beam $40$.}
            \label{fig:40}
          \end{subfigure}
          \vspace{0.5em}
          \begin{subfigure}{\linewidth}
          \centering
            \includegraphics[width=\textwidth]{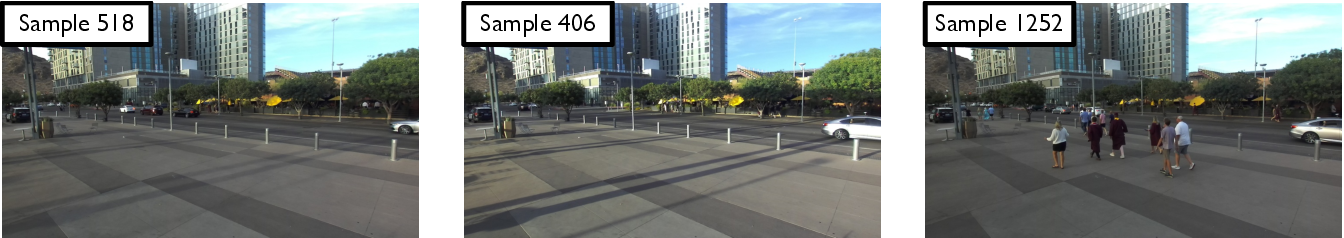}
            \caption{Vision data visualization for the first $3$ samples associated with optimal beam $60$.}
            \label{fig:60}
          \end{subfigure}
          \caption{Vision data visualization for the first $3$ samples associated with optimal beams $20$, $40$, and $60$, respectively.}
          \label{fig:image}
    \end{figure}

    Fig. \ref{fig:image} shows the first three samples when the optimal beam is $20$, $40$, and $60$, respectively. Notice that the UE is the gray car with the sensor at the top, and the optimal beam index gets progressively larger as it travels from the left side of the screen over to the right side. For example, in samples $540$ and $163$, the UE locations are nearly identical, but the weather conditions differ. As shown in Fig. \ref{fig:pow}, this leads to noticeable differences in the received power. From this we can see that the weather condition affects the received power. 

    \begin{figure}[t]
        \centering
        \begin{subfigure}[t]{0.48\columnwidth} 
            \centering
            \includegraphics[width=\textwidth]{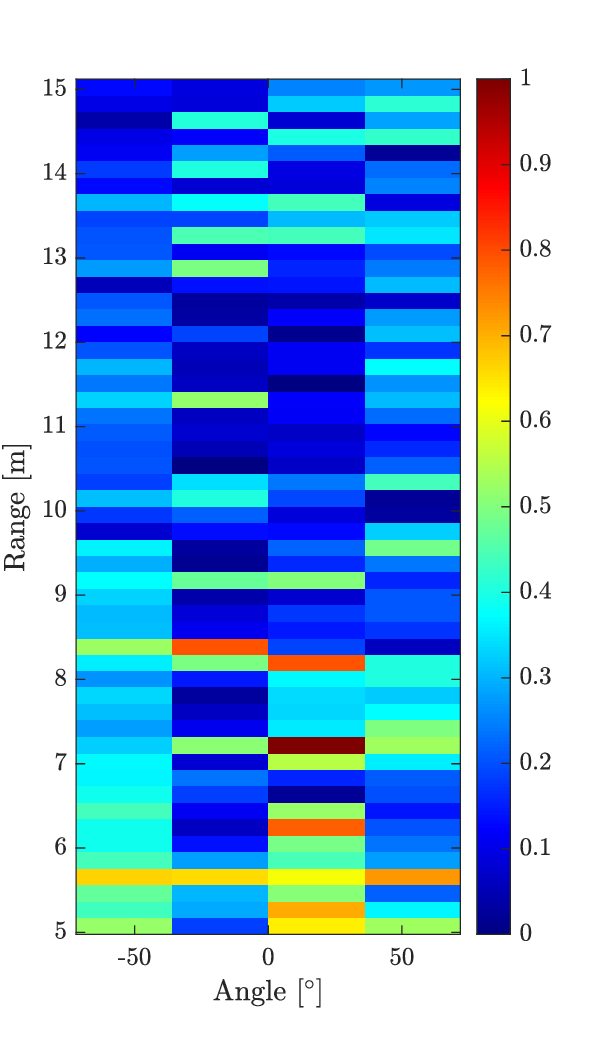} 
            \caption{Range-angle map of sample $86$.}
            \label{fig:ra}
        \end{subfigure}
        \hfill 
        \begin{subfigure}[t]{0.48\columnwidth}
            \centering
            \includegraphics[width=\textwidth]{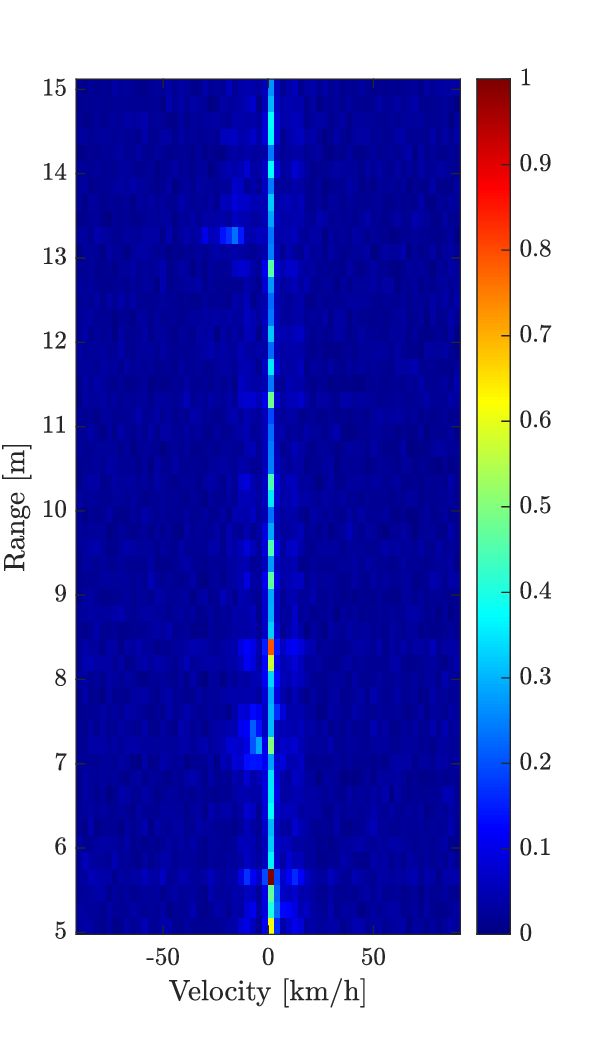} 
            \caption{Range-velocity map of sample $86$.}
            \label{fig:rv}
        \end{subfigure}
        \caption{Radar data visualization of sample $86$.}
        \label{fig:radar}
    \end{figure}

    Fig. \ref{fig:radar} shows two subplots of samlple $86$: (a) the RA map and (b) the range-velocity (RV) map, both of which indicate signal energy in terms of color intensity. The radar system successfully detected multiple objects within the environment. A significant portion of these detects objects appear to be either stationary, functioning as environmental scatterers, or moving very slowly, such as pedestrians. This is clearly substantiated by their distinct positions in the angle-distance spectrum and their strong concentration along the $0$ velocity line in the RV map. Furthermore, the radar system is also capable of identifying distinct moving targets at specific distances. For instance, at a range of approximately $7$ meters, the clear angular and velocity information collectively indicates the presence of a moving target that the radar system is actively tracking.

    \begin{figure}[t]
        \centering
        \includegraphics[width=\linewidth]{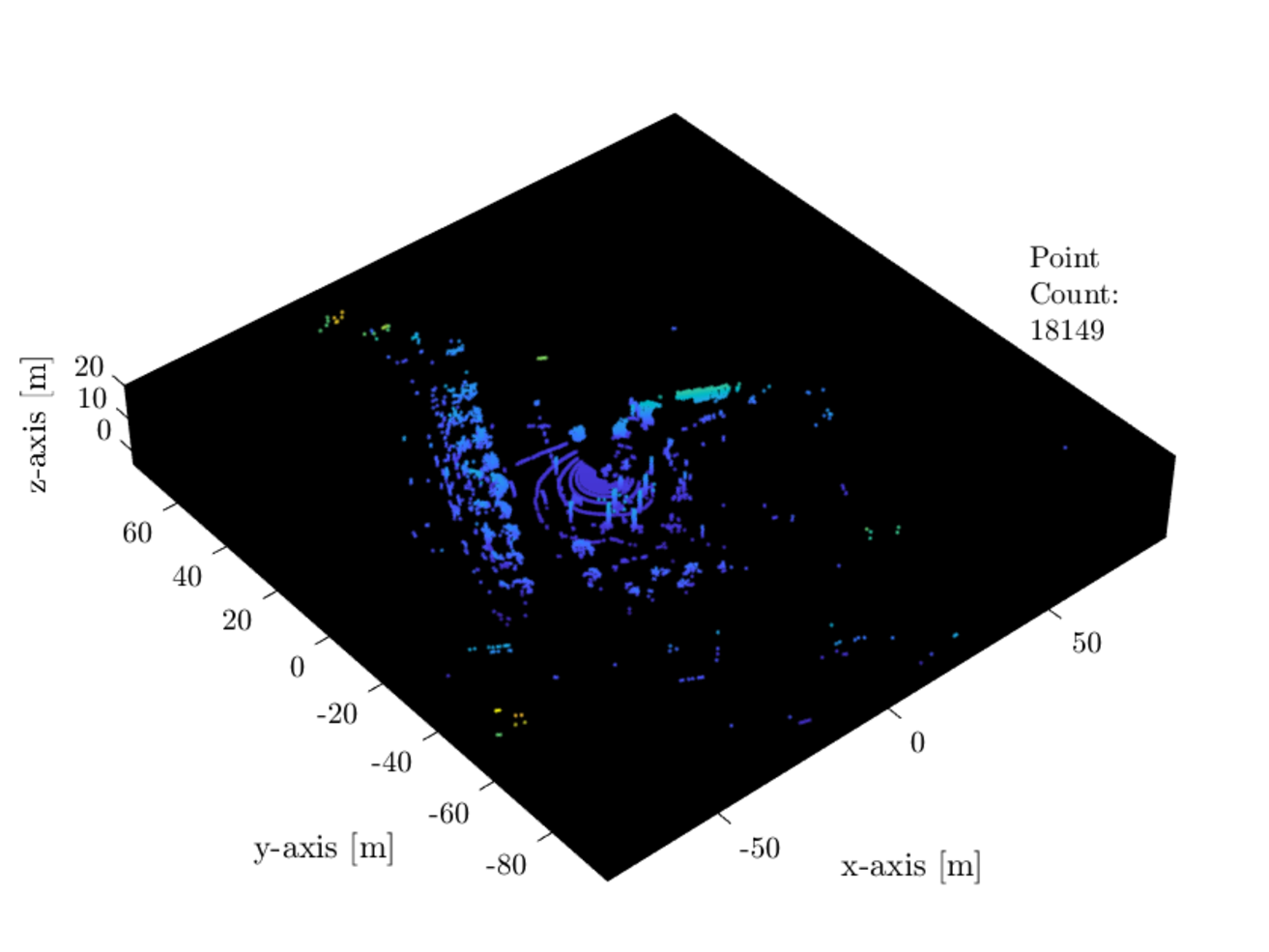}
        \caption{Visualization of LiDAR point cloud data of sample $86$.}
        \label{fig:lidar}
    \end{figure} 

    Fig. \ref{fig:lidar} represents the point cloud data of objects or environments in $3$D space for sample $86$, which, although sparse, reveals the presence of one or more major structures, such as trees, buildings, driveways, etc.

    \begin{figure}[t]
        \centering
        \includegraphics[width=\linewidth]{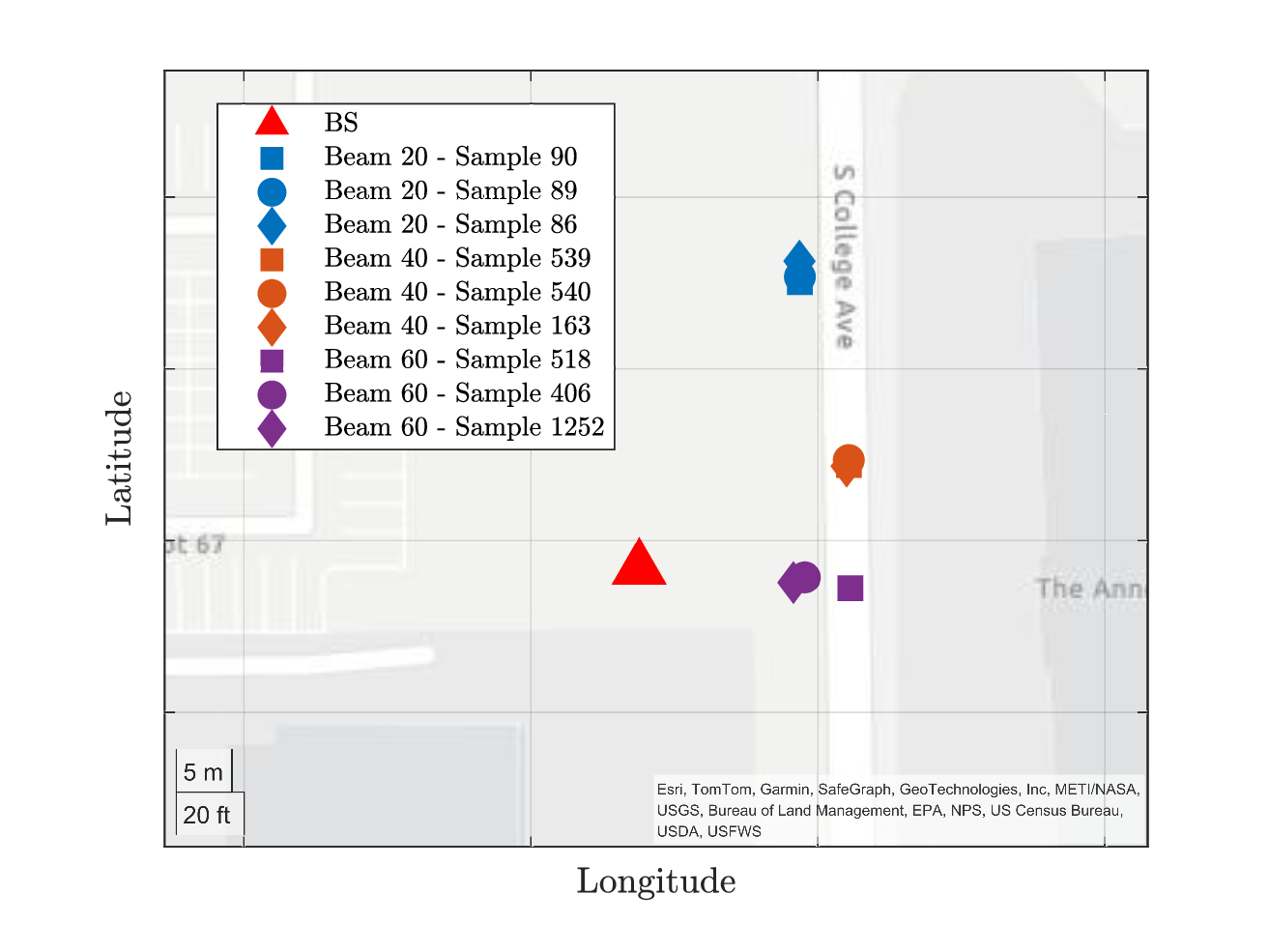}
        \caption{BS and UE GPS location map for the first $3$ samples associated with optimal beams $20$, $40$, and $60$, respectively.}
        \label{fig:gps}
    \end{figure} 

    Fig. \ref{fig:gps} is a geolocation map to show the GPS location of the top three samples with optimal beam indexes of $20$, $40$, and $60$. Combined with the analysis from Fig. \ref{fig:image}, we find that the positional and visual information basically remain aligned.
    
    \subsection{Standard Prediction}

\begin{figure}[t]
        \centering
        \includegraphics[width=\linewidth]{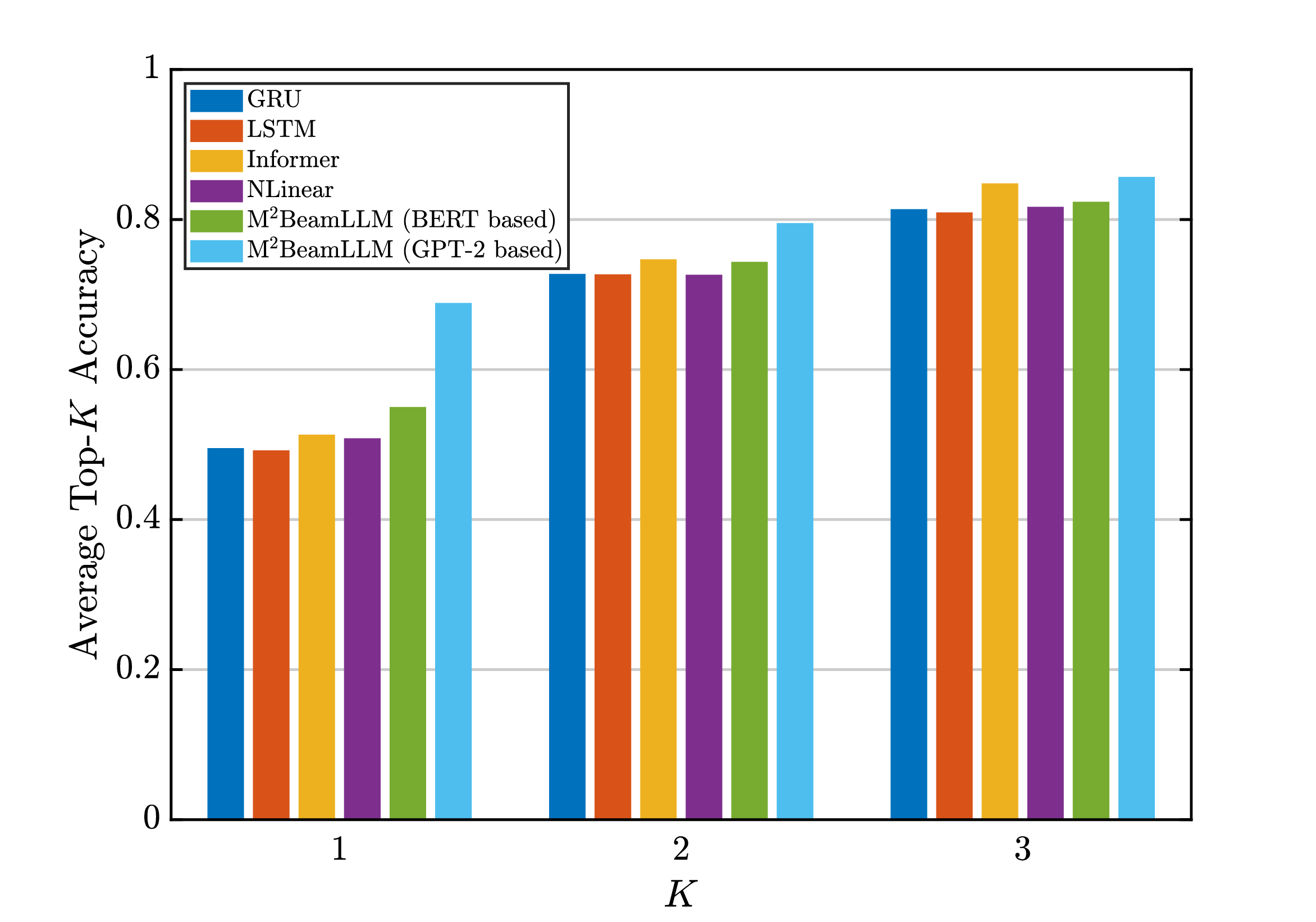}
        \caption{Average Top-$K$ accuracy performance of the proposed method comparing to several baselines in the standard prediction task.}
        \label{fig:topk_std}
    \end{figure}

    Fig. \ref{fig:topk_std} presents a comparative analysis of Top-$1$, Top-$2$, and Top-$3$ accuracy for standard beam predictions across all evaluated models. As expected, increasing the value of $K$ leads to improved accuracy for all models, as the correct label is more likely to appear within the top-$K$ candidates. Despite the large number of parameters, we only activated a small fraction for SFT, so we attribute the high performance to the successful knowledge activation of LLM. Notably, the proposed GPT-$2$-based M\textsuperscript{2}BeamLLM achieves the highest Top-1 accuracy of $\mathbf{68.9\%}$, surpassing the second-best model, BERT-based M\textsuperscript{2}BeamLLM ($55.0\%$), by a substantial $13.9\%$. This significant margin highlights the superior suitability of the transformer decoder architecture (e.g., autoregressive masking in GPT-$2$) for future beam index prediction, compared to the bidirectional encoder architecture used in BERT. Traditional RNN-based models show relatively weak performance at $K=1$, but exhibit notable gains at higher $K$ values. This suggests that while RNNs may struggle to rank the most likely beams correctly, they still offer reasonable overall coverage among top predictions. The Informer model approaches GPT-$2$-based performance at $K=3$, illustrating the strength of its attention mechanism in capturing long-range dependencies. Meanwhile, the simple NLinear model performs comparably to RNNs, reflecting the surprising effectiveness of linear models when trained with sufficient data.

\begin{figure}[t]
        \centering
        \includegraphics[width=\linewidth]{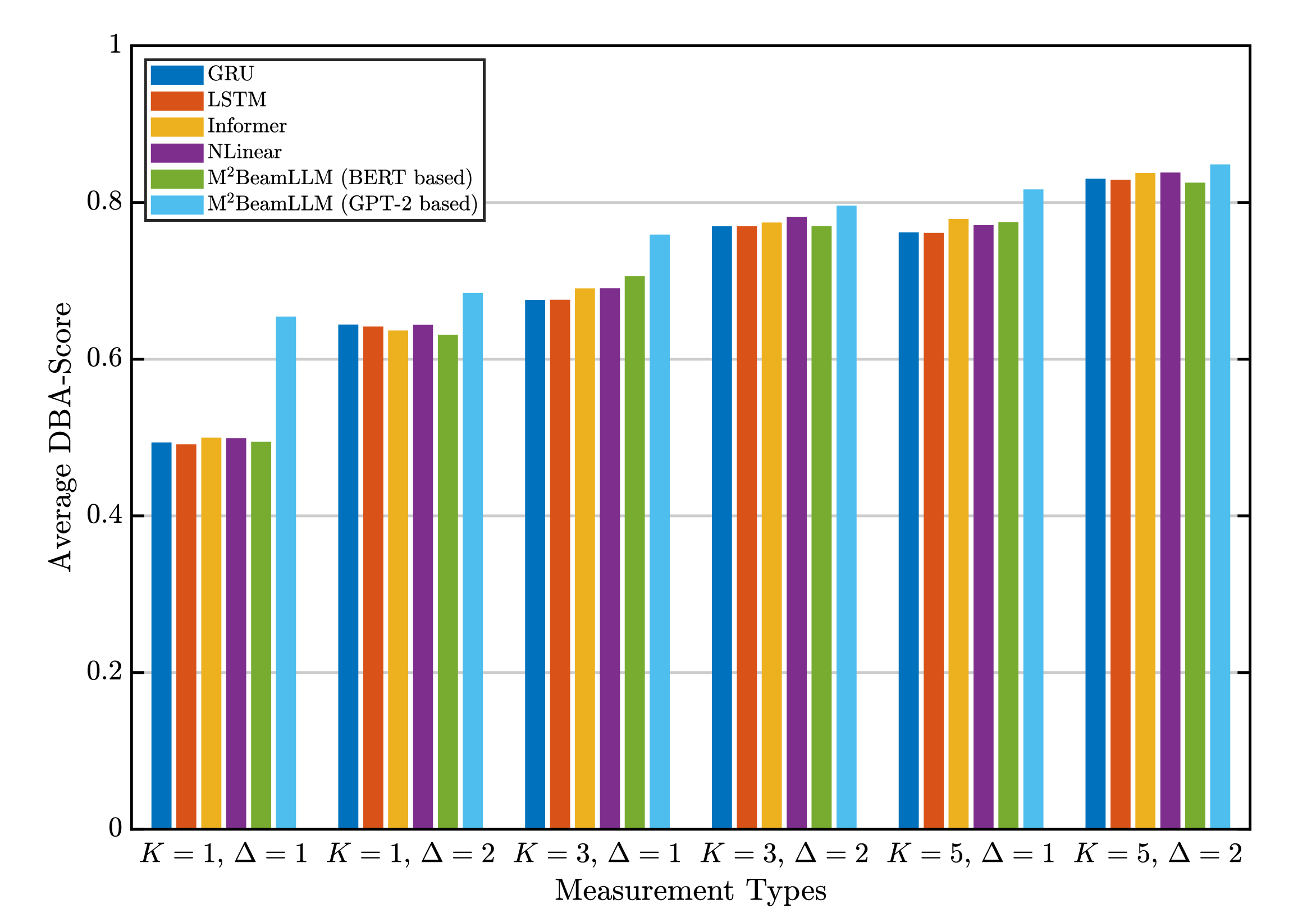}
        \caption{Average DBA-score performance of the proposed method comparing to several baselines in the standard prediction task.}
        \label{fig:dba_std}
    \end{figure}

    Fig. \ref{fig:dba_std} presents the DBA-score as a complementary metric to further differentiate model performance and their temporal alignment behavior under varying tolerance thresholds. As observed, all models demonstrate a progressive improvement in performance as the top-$K$ value increases (from $1$ to $5$) and the tolerance threshold $\Delta$ is relaxed (from $1$ to $2$), indicating greater flexibility in acceptable prediction errors. However, the extent of this improvement is highly architecture-dependent. The proposed GPT-$2$-based M\textsuperscript{2}BeamLLM consistently achieves the highest DBA-scores across all settings, further confirming its robust capability in both beam prediction accuracy. As the tolerance is relaxed to $\Delta=2$, the performance gap between models becomes noticeably smaller. This suggests that a more permissive evaluation criterion reduces the sensitivity to exact temporal alignment, thereby diminishing the relative advantage of more sophisticated models under looser error constraints.

    \subsection{Few-Shot Prediction}
\begin{figure}[t]
        \centering
        \includegraphics[width=\linewidth]{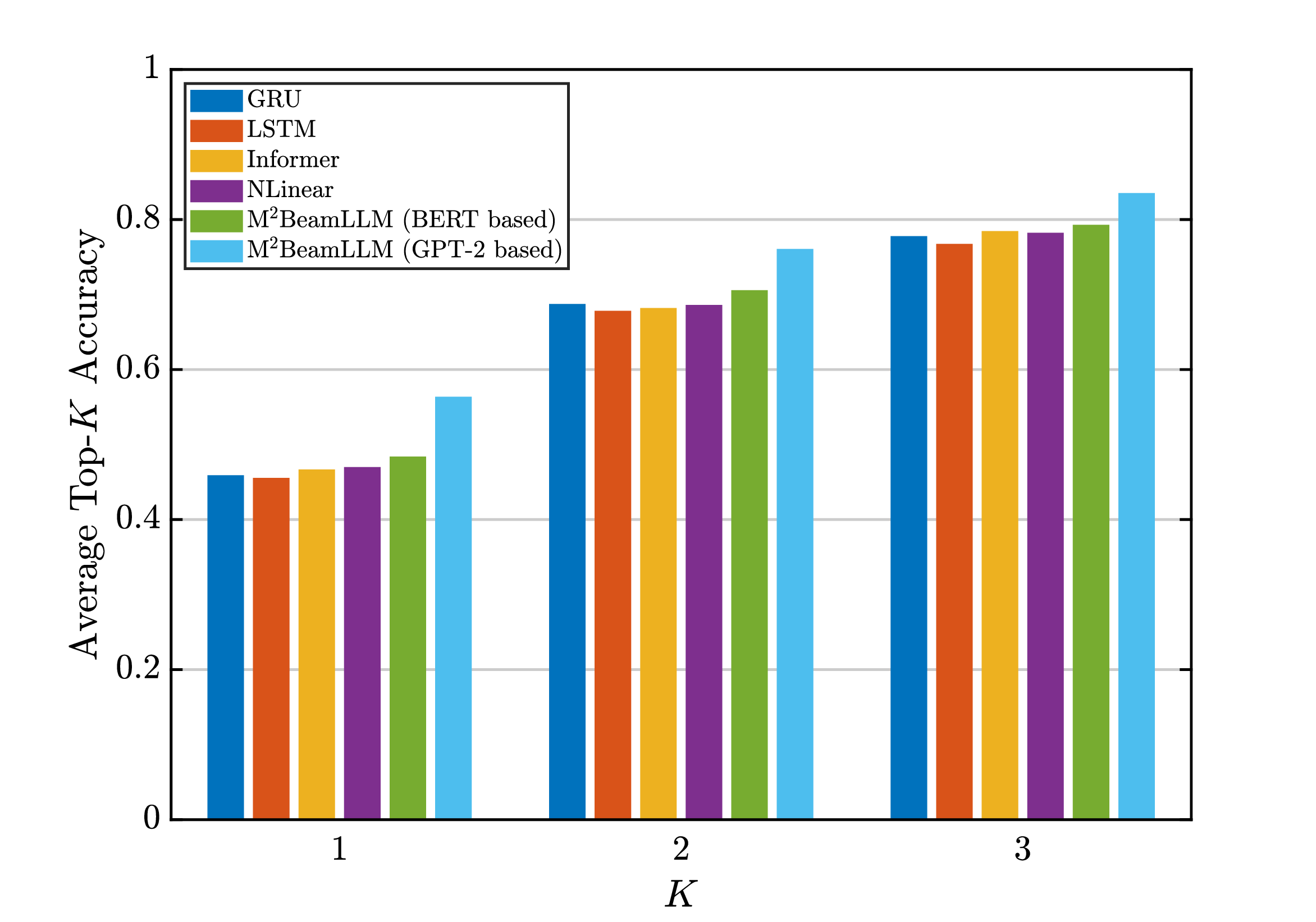}
        \caption{Average Top-$K$ accuracy performance of the proposed method comparing to several baselines in the few-shot prediction task.}
        \label{fig:topk_few}
    \end{figure}

\begin{figure}[t]
        \centering
        \includegraphics[width=\linewidth]{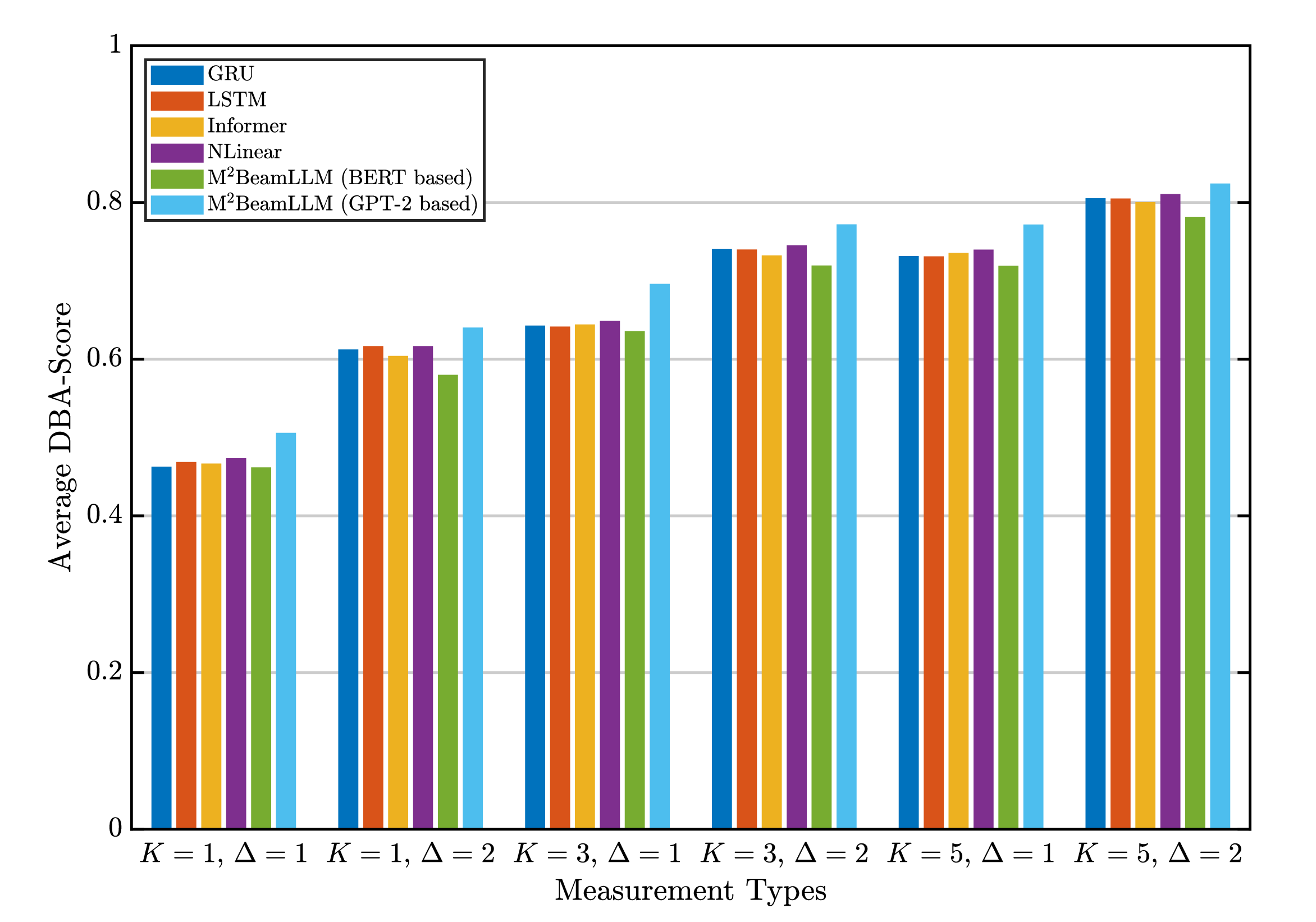}
        \caption{Average DBA-score performance of the proposed method comparing to several baselines in the few-shot prediction task.}
        \label{fig:dba_few}
    \end{figure}

    In Figs. \ref{fig:topk_few} and \ref{fig:dba_few}, we present the top-$K$ accuracy and DBA-score performance for the few-shot prediction task. It is important to note that at this point $H < T$, we need to pad the input of M\textsuperscript{2}BeamLLM with zeros to extend it to $T$ in the time dimension. Despite the overall performance decline, M\textsuperscript{2}BeamLLM continues to exhibit superior performance, maintaining its leading position among the evaluated models. However, the Top-$2$ and Top-$3$ accuracy metrics remain largely unaffected. 

    \subsection{Comparative Analysis and Component Ablation Study}
    In this section, we conduct a series of ablation studies to validate the effectiveness and superiority of the proposed M\textsuperscript{2}BeamLLM framework. Specifically, we design experiments to investigate two key aspects: $(1)$ the impact of different modality combinations on beam prediction performance, and $(2)$ the influence of the number of frozen layers in the pre-trained LLM during supervised fine-tuning.
    
    \subsubsection{Comparative Analysis of Different Modalities}
    \begin{figure}[t]
        \centering
        \includegraphics[width=\linewidth]{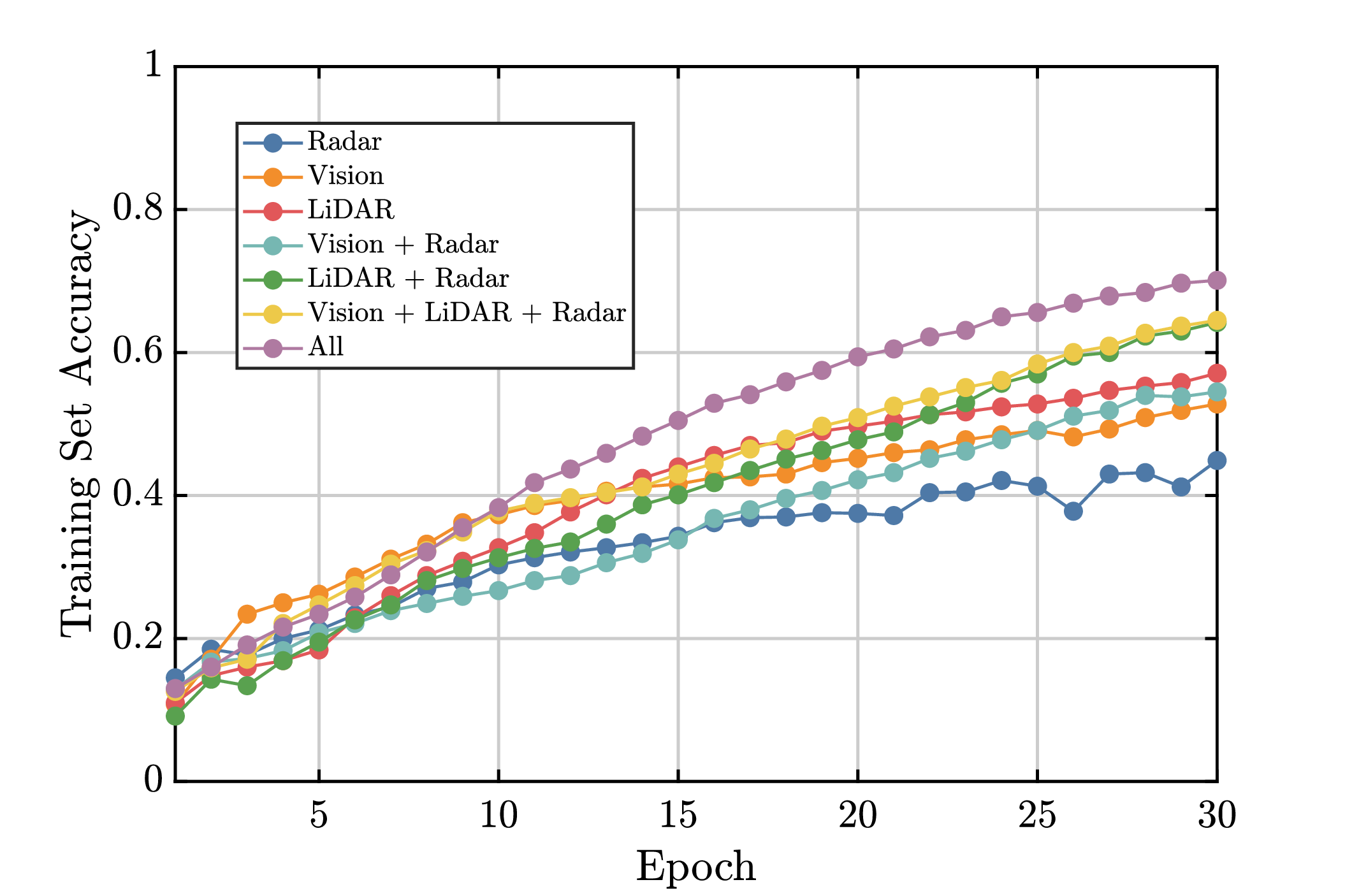}
        \caption{Top-$1$ training accuracy performance comparison of different combinations of sensing data.}
        \label{fig:modalTraining}
    \end{figure} 

    \begin{figure}[t]
        \centering
        \includegraphics[width=\linewidth]{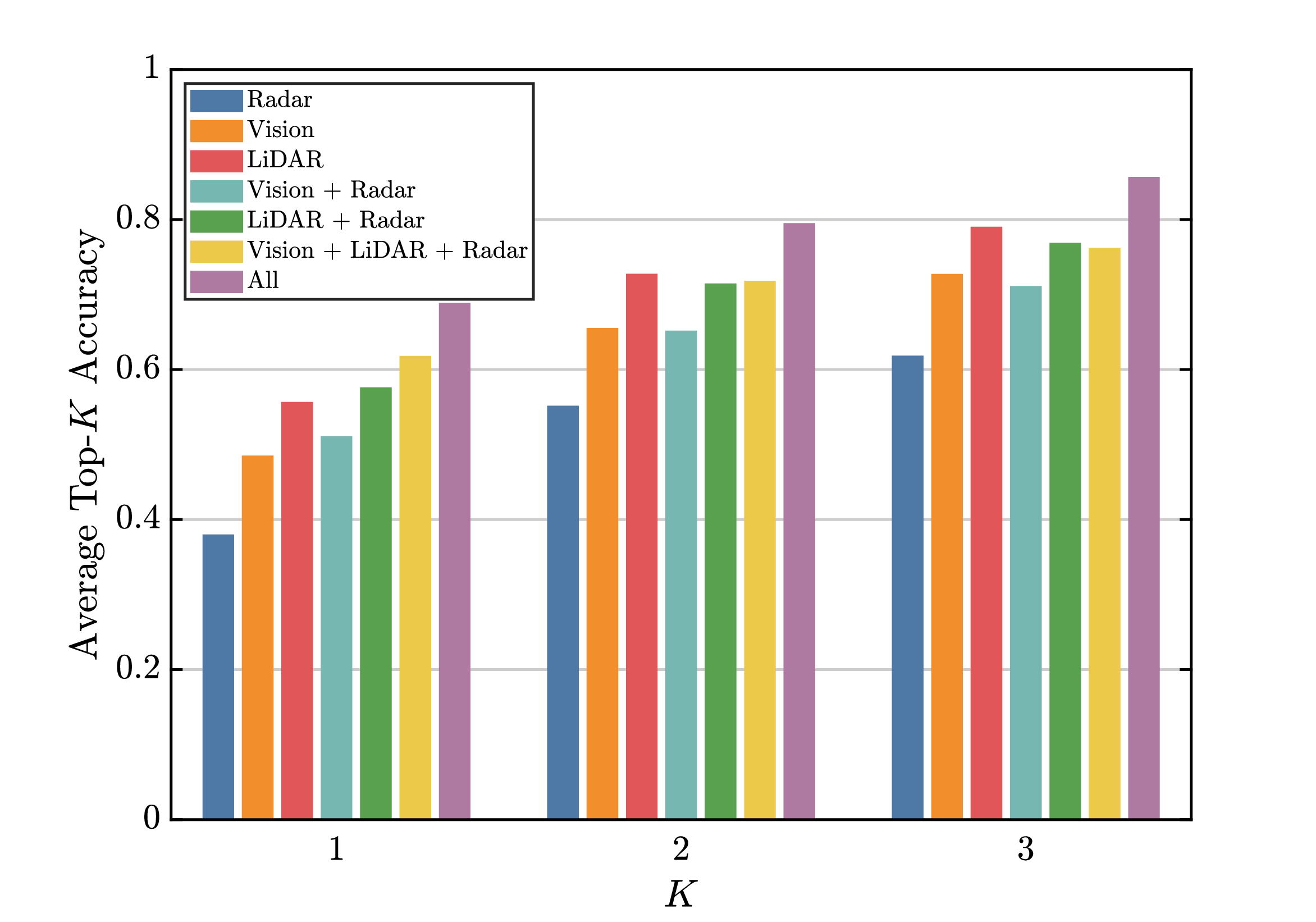}
        \caption{Average Top-$K$ accuracy performance of different combinations of sensing data.}
        \label{fig:modalTopK}
    \end{figure} 

    \begin{figure}[t]
        \centering
        \includegraphics[width=\linewidth]{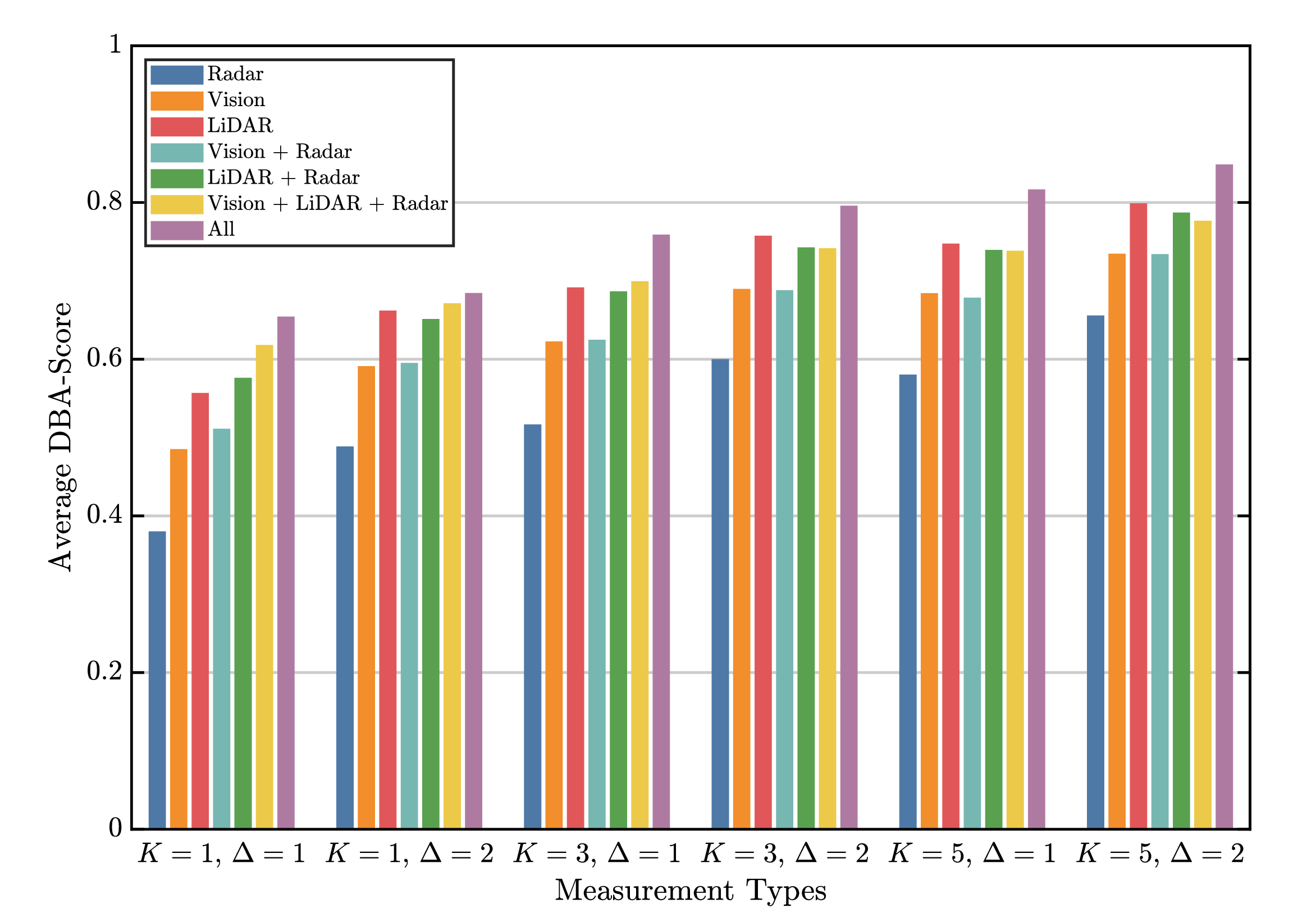}
        \caption{Average DBA-score performance of different combinations of sensing data.}
        \label{fig:modalDBA}
    \end{figure} 
    
    We focus on evaluating the performance impact of different combinations of sensing modalities. This analysis investigates how varying the number of input modalities influences beam prediction accuracy. The training and evaluation results are presented in Figs. \ref{fig:modalTraining}, \ref{fig:modalTopK}, and \ref{fig:modalDBA}, showcasing performance across three single-modality, two dual-modality, one tri-modality, and one full-modality configurations. A clear trend of performance enhancement through multimodal fusion emerges, where an increased number of modalities generally leads to improved beam prediction accuracy. For instance, in terms of Top-1 accuracy, the combination of vision and radar modalities achieves an accuracy of $51.1\%$, representing an improvement of $2.6\%$ and $13.1\%$ over using vision-only and radar-only modalities, respectively. Further incorporating the LiDAR modality boosts the accuracy by an additional $10.7\%$. Building upon this improvement, the inclusion of GPS data yields a further $7.1\%$ gain in accuracy. These results highlight the complementary benefits of multimodal sensing and validate the effectiveness of the proposed M\textsuperscript{2}BeamLLM framework in leveraging diverse sensing inputs for robust and accurate beam prediction.
    
    \subsubsection{Comparative Analysis of the Performance of Different Frozen Pre-training Layers}
    \begin{figure}[t]
        \centering
        \includegraphics[width=\linewidth]{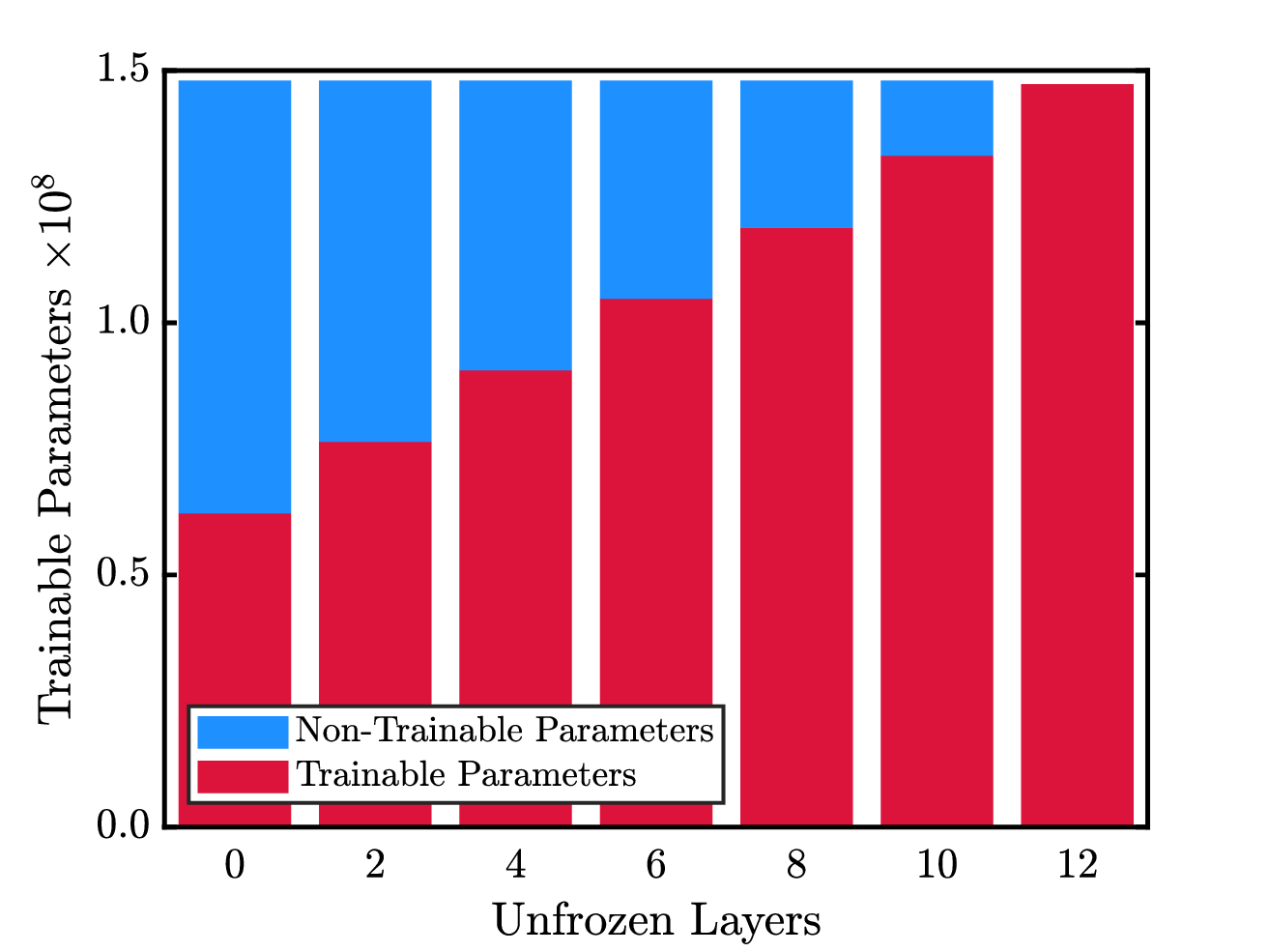}
        \caption{Distribution of trainable versus non-trainable parameters across unfrozen layers.}
        \label{fig:frozenParams}
    \end{figure} 
    \begin{figure}[t]
        \centering
        \includegraphics[width=\linewidth]{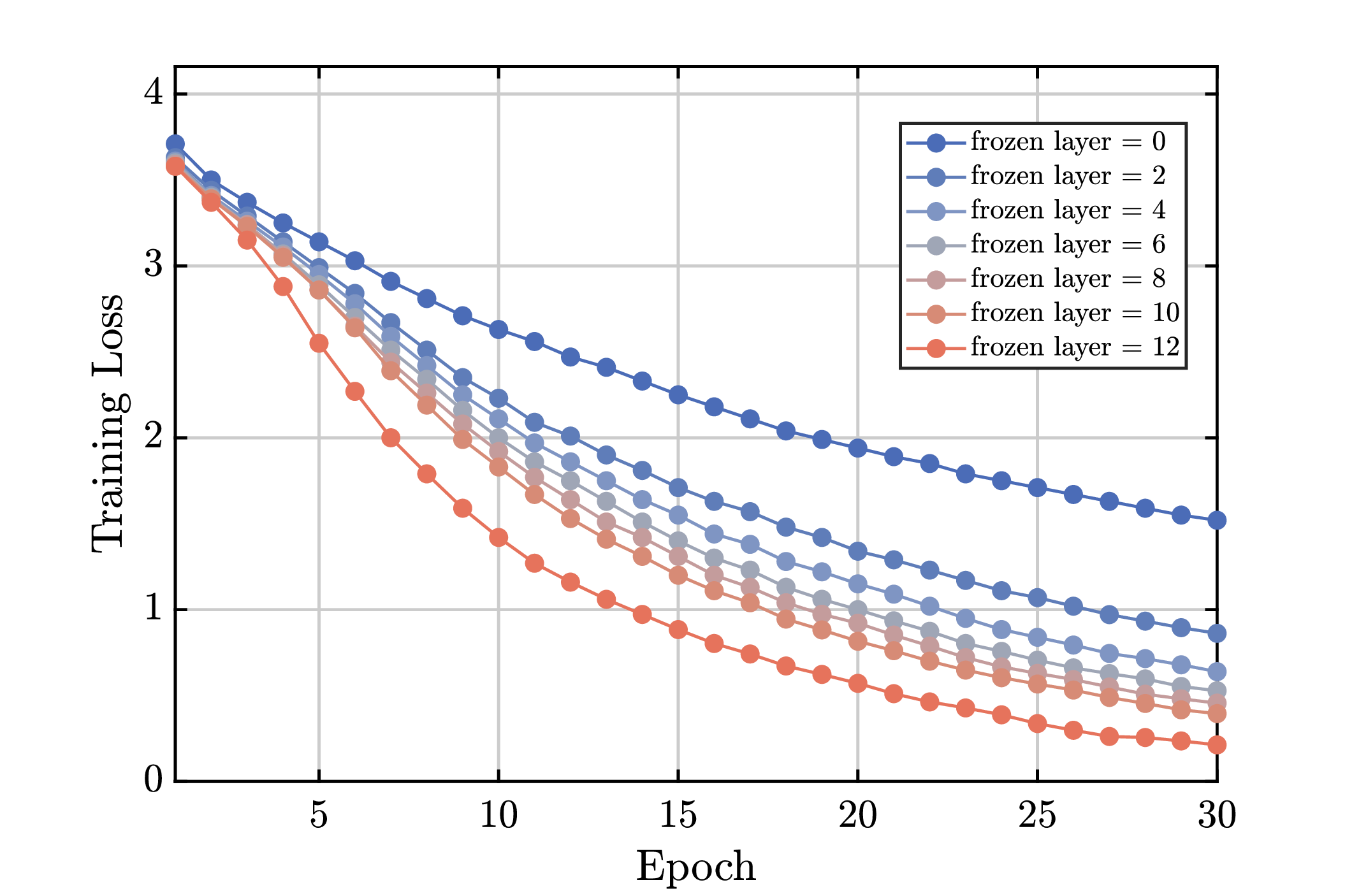}
        \caption{Effect of the number of frozen layers on training loss.}
        \label{fig:frozenTraining}
    \end{figure} 
    In this section, we explore the impact of unfreezing different numbers of layers in the LLM backbone model on training and performance.

    We observe that as more transformer layers are unfrozen, the number of trainable parameters increases from $62.3$ M ($0$ layers) to $147.3$ M ($12$ layers), with the final validation loss decreasing from $1.52$ to $0.21$ after $30$ epochs. Top-$K$ accuracies and DBA-scores consistently improve with deeper fine-tuning. For instance, Top-$1$ accuracy increases from $47.4\%$ ($0$ layers) to $\mathbf{85.7}\%$ ($12$ layers), while DBA-score with Top-$1$, $\Delta=1$ reaches $\mathbf{0.94}$ at full fine-tuning. These results demonstrate that deeper fine-tuning not only improves convergence but also enhances semantic and temporal alignment of predictions.

    Furthermore, even partial fine-tuning provides substantial gains: unfreezing just $4$ layers results in significant improvements across all metrics (e.g., Top-$1$ accuracy from $47.4\%$ $\to$ $72.4\%$, DBA-score with Top-$1$, $\Delta=1$ from $0.47$ $\to$ $0.72$). Deeper tuning beyond $8$ layers continues to yield improvements, though with diminishing returns relative to computational cost.

    If computational resources allow, we recommend full fine-tuning ($12$ layers) to achieve optimal performance. For resource-constrained scenarios (e.g., limited GPU memory or training time), unfreezing $6–8$ layers strikes a highly cost-effective balance, achieving Top-$3$ accuracy around $90\%$, with significantly reduced training overhead.

    \begin{figure}[t]
        \centering
        \includegraphics[width=\linewidth]{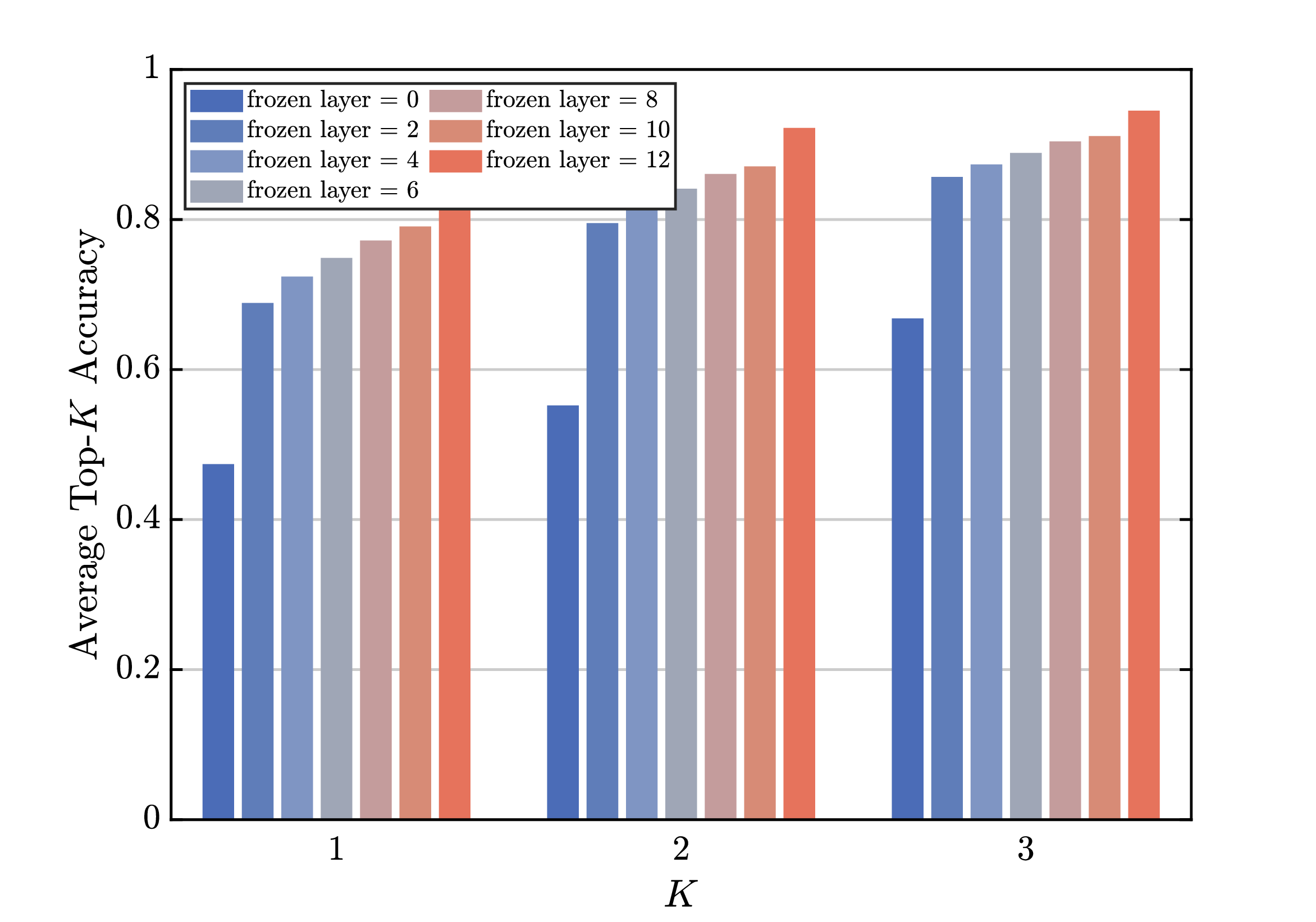}
        \caption{Comparison of average Top-$K$ accuracy for different numbers of frozen layers.}
        \label{fig:frozenTpK}
    \end{figure} 

    \begin{figure}[t]
        \centering
        \includegraphics[width=\linewidth]{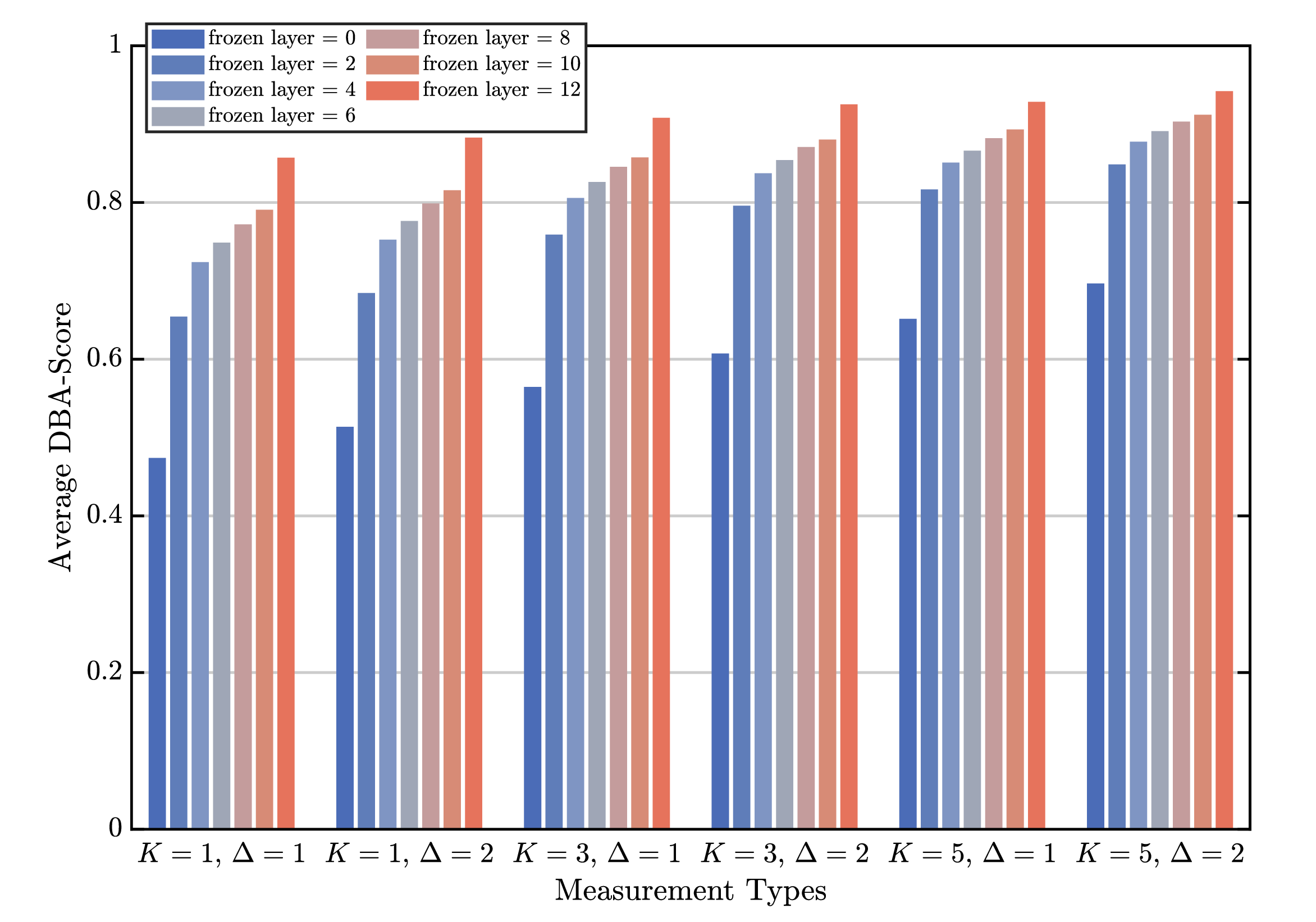}
        \caption{Comparison of average DBA-score accuracy for different numbers of frozen layers.}
        \label{fig:frozenDBA}
    \end{figure} 

    \subsection{Complexity Analysis}

    To evaluate the feasibility of deploying the proposed method in practical scenarios, we conduct a comparative analysis of the model's number of parameters and inference cost against various baseline approaches. This assessment provides insights into the computational demands and potential deployment challenges associated with each model. The results of this comparison are presented in Table \ref{tab:complexity}. 
    
    Overall, the inference time per sample for all models is shorter than the data sampling and synchronization intervals.\footnote{It is noted that the synchronization interval is $100$ ms for DeepSense $6$G Scenario $32$ (https://www.deepsense6g.net/data-collection/).} Notably, due to the inference acceleration capabilities inherent in GPT and BERT models, M\textsuperscript{2}BeamLLM exhibits significantly reduced inference time compared to models like Informer and NLinear. Consequently, the proposed M\textsuperscript{2}BeamLLM demonstrates potential for real-time beam prediction services.

    \begin{table}[t]
        \centering
        \caption{Comparison of the network parameters and the interference cost per test sample.}
        \small
        \setlength{\tabcolsep}{6pt}
        \renewcommand{\arraystretch}{1.2} 
        \begin{tabularx}{\linewidth}{l X c c}
            \toprule
            \textbf{Category} & \textbf{Model} 
            & \begin{tabular}[c]{@{}c@{}}\textbf{Number of} \\ \textbf{Parameters}\end{tabular} 
            & \begin{tabular}[c]{@{}c@{}}\textbf{Inference} \\ \textbf{Time ($10^{-4}$ s)}\end{tabular} \\
            \midrule
            \multirow{4}{*}{Encoders} 
                & Vision                & $11,348,992$ & / \\
                & Radar                 & $175,254$ & / \\
                & LiDAR                 & $11,203,072$ & /  \\
                & GPS                   & $6688$ & / \\
                & \textbf{Total}                 & $22,734,006$ & / \\
            \midrule
            \multirow{6}{*}{Prediction}
                & GRU                           & $23$ M & $1.34$ \\
                & LSTM                          & $24$ M & $1.44$ \\
                & Informer                      & $27$ M & $4.20$ \\
                & NLinear                       & $23$ M & $3.68$ \\
                & M\textsuperscript{2}BeamLLM (GPT-2)  & $147$ M & $2.55$ \\
                & M\textsuperscript{2}BeamLLM (BERT)   & $132$ M & $1.84$ \\
            \bottomrule
        \end{tabularx}
        \label{tab:complexity}
    \end{table}

\section{Conclusion}
    This paper introduced M\textsuperscript{2}BeamLLM, a neural network framework for beam prediction in mmWave mMIMO communication systems. The framework integrates multimodal sensing data, including images, radar, LiDAR, and GPS, and harnesses the powerful inference capabilities of LLMs. M\textsuperscript{2}BeamLLM utilizes multimodal alignment and fusion in conjunction with SFT to improve the accuracy and robustness of beam prediction, significantly outperforming traditional models in both standard and few-shot scenarios. Our findings confirm that the prediction performance of multimodal combinations adheres to the general principle that greater diversity leads to better outcomes. Additionally, we observed that increasing the number of fine-tuning layers enhances prediction performance, albeit at the cost of higher training overhead. Moreover, the inference speed of M\textsuperscript{2}BeamLLM remains comparable to or even faster than baseline models, despite its parameter count being several times higher. This study provided an efficient and intelligent beam prediction solution for V$2$I mmWave mMIMO communication systems.
  
    Nonetheless, there are potential directions for further enhancing this work. A promising avenue involves the design and initialization of a novel large model from the ground up, rather than relying on existing pre-trained LLMs. This foundational model would undergo rigorous pre-training on an extensive dataset, with subsequent fine-tuning adjustments applied during its deployment phase to fully harness its inherent generalization capabilities. Building on this, a more advanced approach would be to design such agent specifically tailored for wireless communication tasks.  It would integrate diverse communication and sensing information, such as CSI, images, radar, LiDAR, and GPS data, collected across various times, frequencies, and spatial contexts. Subsequently, at the decoding stage, different decoder modules can be configured to transform the model into an agent framework. This approach significantly enriches the variety of downstream tasks the model can handle-such as beam prediction, channel prediction, blockage prediction, and power control—and allows for direct invocation of different functionalities based on specific task requirements. Moreover, recognizing the practical challenges associated with supervised learning reliant on online labeling, reinforcement learning (RL) presents a viable solution for these tasks \cite{rl}. Capitalizing on the formidable inference power of LLMs, this methodology is expected to facilitate the acquisition of underlying regularities within sensing data and beam predictions, entirely bypassing the requirement for explicit labeled data.


\balance
\bibliographystyle{IEEEtran}
\bibliography{IEEEabrv, ref}

\begin{thebibliography}{10}
\providecommand{\url}[1]{#1}
\csname url@samestyle\endcsname
\providecommand{\newblock}{\relax}
\providecommand{\bibinfo}[2]{#2}
\providecommand{\BIBentrySTDinterwordspacing}{\spaceskip=0pt\relax}
\providecommand{\BIBentryALTinterwordstretchfactor}{4}
\providecommand{\BIBentryALTinterwordspacing}{\spaceskip=\fontdimen2\font plus
\BIBentryALTinterwordstretchfactor\fontdimen3\font minus \fontdimen4\font\relax}
\providecommand{\BIBforeignlanguage}[2]{{%
\expandafter\ifx\csname l@#1\endcsname\relax
\typeout{** WARNING: IEEEtran.bst: No hyphenation pattern has been}%
\typeout{** loaded for the language `#1'. Using the pattern for}%
\typeout{** the default language instead.}%
\else
\language=\csname l@#1\endcsname
\fi
#2}}
\providecommand{\BIBdecl}{\relax}
\BIBdecl

\bibitem{mmWave}
M.~Xiao, S.~Mumtaz, Y.~Huang, L.~Dai, Y.~Li, M.~Matthaiou, G.~K. Karagiannidis, E.~Bjornson, K.~Yang, C.-L. I, and et~al., ``Millimeter wave communications for future mobile networks,'' \emph{{IEEE} J. Sel. Areas Commun.}, vol.~35, no.~9, p. 1909–1935, Sep 2017.

\bibitem{ml}
Q.~Xue, C.~Ji, S.~Ma, J.~Guo, Y.~Xu, Q.~Chen, and W.~Zhang, ``A survey of beam management for mmwave and thz communications towards 6g,'' \emph{{IEEE} Commun. Surveys Tuts.}, vol.~26, no.~3, pp. 1520--1559, 2024.

\bibitem{csi}
M.~Alrabeiah and A.~Alkhateeb, ``Deep learning for mmwave beam and blockage prediction using sub-6 ghz channels,'' \emph{{IEEE} Trans. Commun.}, vol.~68, no.~9, pp. 5504--5518, 2020.

\bibitem{CV}
S.~Jiang and A.~Alkhateeb, ``{Computer vision aided beam tracking in A real-world millimeter wave deployment},'' in \emph{Proc. IEEE Globecom Workshops (GC Wkshps)}, 2022, pp. 142--147.

\bibitem{Radar}
U.~Demirhan and A.~Alkhateeb, ``{Radar aided 6G beam prediction: Deep learning algorithms and real-world demonstration},'' in \emph{Proc. IEEE Wireless Communications and Networking Conference (WCNC)}, 2022, pp. 2655--2660.

\bibitem{LiDAR}
S.~Jiang, G.~Charan, and A.~Alkhateeb, ``{LiDAR aided future beam prediction in real-world millimeter wave V2I communications},'' \emph{IEEE Wireless Commun. Lett.}, vol.~12, no.~2, pp. 212--216, 2023.

\bibitem{GPS}
J.~Morais, A.~Bchboodi, H.~Pezeshki, and A.~Alkhateeb, ``{Position-aided beam prediction in the real world: How useful GPS locations actually are?}'' in \emph{Proc. IEEE International Conference on Communications (ICC)}, 2023, pp. 1824--1829.

\bibitem{GPT-4}
OpenAI, ``{GPT-4} technical report,'' Accessed Mar. 2023. [Online]. Available: https://arxiv.org/abs/2303.08774.

\bibitem{deepseek}
D.~Guo, D.~Yang, H.~Zhang, J.~Song, R.~Zhang, R.~Xu, Q.~Zhu, S.~Ma, P.~Wang, X.~Bi \emph{et~al.}, ``{Deepseek-r1: Incentivizing reasoning capability in LLMs via reinforcement learning},'' \emph{arXiv preprint arXiv:2501.12948}, 2025.

\bibitem{BeamLLM}
C.~Zheng, J.~He, G.~Cai, Z.~Yu, and C.~G. Kang, ``{BeamLLM: Vision-empowered mmWave beam prediction with large language models},'' \emph{arXiv preprint arXiv:2503.10432}, 2025.

\bibitem{CV+GPS}
L.~Cheng, H.~Zhang, B.~Di, D.~Niyato, and L.~Song, ``{Large language models empower multimodal integrated sensing and communication},'' \emph{{IEEE} Commun. Mag.}, vol.~63, no.~5, pp. 190--197, 2025.

\bibitem{Multi_modal}
B.~Shi, M.~Li, M.-M. Zhao, M.~Lei, and L.~Li, ``{Multimodal deep learning empowered millimeter-wave beam prediction},'' in \emph{Proc. IEEE 99th Vehicular Technology Conference (VTC2024-Spring)}, 2024, pp. 1--6.

\bibitem{MoE}
K.~Zhang, W.~Yu, H.~He, S.~Song, J.~Zhang, and K.~B. Letaief, ``{Multimodal deep learning-empowered beam prediction in future THz ISAC systems},'' \emph{arXiv preprint arXiv:2505.02381}, 2025.

\bibitem{Transf}
Y.~Cui, J.~Nie, X.~Cao, T.~Yu, J.~Zou, J.~Mu, and X.~Jing, ``{Sensing-assisted high reliable communication: A transformer-based beamforming Approach},'' \emph{{IEEE} J. Sel. Topics Signal Process.}, vol.~18, no.~5, pp. 782--795, 2024.

\bibitem{SFT}
C.~Zhou, P.~Liu, P.~Xu, S.~Iyer, J.~Sun, Y.~Mao, X.~Ma, A.~Efrat, P.~Yu, L.~Yu, S.~Zhang, G.~Ghosh, M.~Lewis, L.~Zettlemoyer, and O.~Levy, ``{LIMA: Less is more for alignment},'' in \emph{Proc. Neural Information Processing Systems (NIPS)}, 2023.

\bibitem{ImageNet}
J.~Deng, W.~Dong, R.~Socher, L.-J. Li, K.~Li, and L.~Fei-Fei, ``{ImageNet: A large-scale fierarchical image database},'' in \emph{Proc. IEEE Conference on Computer Vision and Pattern Recognition (CVPR)}, Jun. 2009, pp. 248--255.

\bibitem{ResNet}
K.~He, X.~Zhang, S.~Ren, and J.~Sun, ``{Deep residual learning for image recognition},'' in \emph{Proc. IEEE Conference on Computer Vision and Pattern Recognition (CVPR)}, Jun. 2016, pp. 770--778.

\bibitem{ZP_FFT}
Q.~Guoqing, ``{High accuracy range estimation of FMCW level radar based on the phase of the zero-padded FFT},'' in \emph{Proc. 7th International Conference on Signal Processing (ICSP)}, vol.~3, 2004, pp. 2078--2081.

\bibitem{LeNet}
Y.~Lecun, L.~Bottou, Y.~Bengio, and P.~Haffner, ``{Gradient-based learning applied to document recognition},'' \emph{Proc. {IEEE}}, vol.~86, no.~11, pp. 2278--2324, 1998.

\bibitem{Alignment}
S.~Li and H.~Tang, ``{Multimodal alignment and fusion: A survey},'' \emph{arXiv preprint arXiv:2411.17040}, 2024.

\bibitem{CLIP}
A.~Radford, J.~W. Kim, C.~Hallacy, A.~Ramesh, G.~Goh, S.~Agarwal, G.~Sastry, A.~Askell, P.~Mishkin, J.~Clark \emph{et~al.}, ``{Learning transferable visual models from natural language supervision},'' in \emph{Proc. International Conference on Machine Learning (ICML)}, Jul. 2021, pp. 8748--8763.

\bibitem{DeepSense6G}
A.~Alkhateeb, G.~Charan, T.~Osman, A.~Hredzak, J.~Morais, U.~Demirhan, and N.~Srinivas, ``{DeepSense 6G: A large-scale real-world multi-modal sensing and communication dataset},'' \emph{{IEEE} Commun. Mag.}, vol.~61, no.~9, pp. 122--128, 2023.

\bibitem{GPT2}
A.~Radford \emph{et~al.}, ``{Language models are unsupervised multitask learners},'' \emph{OpenAI blog}, vol.~1, no.~8, p.~9, Feb. 2019.

\bibitem{GRU}
K.~Cho, B.~Van~Merri{\"e}nboer, D.~Bahdanau, and Y.~Bengio, ``{On the properties of neural machine translation: encoder-decoder approaches},'' \emph{arXiv preprint arXiv:1409.1259}, 2014.

\bibitem{LSTM}
S.~Hochreiter and J.~Schmidhuber, ``{Long short-term memory},'' \emph{Neural Computation}, vol.~9, no.~8, pp. 1735--1780, 1997.

\bibitem{NLinear}
A.~Zeng, M.~Chen, L.~Zhang, and Q.~Xu, ``Are transformers effective for time series forecasting?'' 2023.

\bibitem{Informer}
H.~Zhou, S.~Zhang, J.~Peng, S.~Zhang, J.~Li, H.~Xiong, and W.~Zhang, ``{Informer: Beyond efficient transformer for long sequence time-series Forecasting},'' in \emph{Proc. AAAI Conference on Artificial Intelligence}, vol.~35, May. 2021, pp. 11\,106--11\,115.

\bibitem{ADAM}
D.~P. Kingma and J.~Ba, ``{Adam: A method for stochastic optimization},'' in \emph{Proc. International Conference on Learning Representations (ICLR)}, May. 2015.

\bibitem{rl}
M.~Ghassemi, H.~Zhang, A.~Afana, A.~B. Sediq, and M.~Erol-Kantarci, ``Multi-modal transformer and reinforcement learning-based beam management,'' \emph{{IEEE} Netw. Lett.}, vol.~6, no.~4, pp. 222--226, 2024.

\end{thebibliography}


\end{document}